\documentclass[a4paper,fleqn]{cas-dc}

\usepackage[numbers]{natbib}

\newcommand{\student}{\mathcal{S}}

\usepackage{algorithm}
\usepackage{algpseudocode}

\usepackage{amsthm}
\usepackage{bm} 

\usepackage{caption,subcaption} 
\usepackage{tabularx}
\usepackage{afterpage}
\usepackage{float} 
\usepackage[section]{placeins} 
\usepackage{adjustbox} 
\usepackage{rotating} 

\begin{document}
\let\WriteBookmarks\relax
\def\floatpagepagefraction{1}
\def\textpagefraction{.001}

\shorttitle{CT-Lite: Learning from Compressed CT}

\shortauthors{S. Yousuf et~al.}

\title[mode = title]{Learning from Compressed CT: Feature Attention Style Transfer and Structured Factorized Projections for Resource-Efficient Medical Image Analysis}

\author[1]{Shadid Yousuf}

\author[1]{S.M. Mahbubur Rahman}

\author[1]{Mohammed Imamul Hassan Bhuiyan}

\affiliation[1]{organization={Department of Electrical and Electronic Engineering, Bangladesh University of Engineering and Technology},
            city={Dhaka},
            country={Bangladesh}}

\RenewDocumentCommand{\printorcid}{}{}

\begin{abstract}
    The deployment of artificial intelligence in medical imaging is hindered by high computational complexity and resource-intensive processing of volumetric data. Although chest computed tomography (CT) volumes offer rich diagnostic information, their use in AI-based diagnosis remains limited due to the computational burden of processing uncompressed volumetric images (typically stored in NIfTI or DICOM format). Addressing the growing need for low-resource deployment and efficient electronic data transfer, we conduct the first systematic study of compressed-CT diagnostic learning by investigating the utilization of JPEG-compressed chest CT volumes for thoracic abnormality detection. We propose Feature Attention Style Transfer (FAST), a novel distillation framework that transfers both activation patterns and structural relationships from high-fidelity CT representations to a spatiotemporal visual encoder operating on compressed inputs. By combining Gram-matrix-based attention style preservation with dual-attention feature alignment, FAST enables robust feature extraction from degraded volumes. Furthermore, we introduce Structured Factorized Projection (SFP), leveraging Block Tensor Train decomposition as a parameter-efficient alternative to dense projection layers, reducing projection-head parameters by almost half. Our contrastive learning pipeline, CT-Lite, integrates these components with a SigLIP-based multimodal alignment objective. Experiments on CT-RATE, NIDCH, and Rad-ChestCT demonstrate that CT-Lite achieves AUROC within 5-7\% of the uncompressed-input baseline across all three datasets, despite operating on compressed inputs with significantly fewer parameters, paving the way for AI-based clinical evaluation under resource constraints.
\end{abstract}



\maketitle

\section{Introduction}

The clinical deployment of AI-based chest computed tomography (CT) analysis is hindered by two notable challenges: the computational cost of processing high-dimensional volumetric data, and the bandwidth required to transmit it. Standard volumetric formats such as DICOM and NIfTI are inherently bandwidth-intensive as they store 12 to 16 bits of grayscale data, which makes them poorly suited to telemedicine and electronic data sharing. These constraints have become more pressing as global radiologist shortages \cite{konstantinidis2024shortage,mirak2025growing,jeganathan2023growing} have driven a push toward AI-assisted radiological interpretation \cite{achour2025role,jing2025ai,slanetz2023promise}. State-of-the-art chest CT foundation models like CT-CLIP \cite{hamamci2026generalist}, COLIPRI \cite{wald2026_colipri} or fVLM \cite{shui2025large} that could meet this demand are themselves resource-intensive, requiring high-end GPU infrastructure for both training and inference, which places them out of reach for most resource-constrained clinical environments. The visual projection head of CT-CLIP alone, for example, contains roughly one billion parameters on its own \cite{hamamci2026generalist}, illustrating how the dense layers that embed volumetric features into a shared multimodal latent space can dominate the overall computational burden.

A natural way to address the bandwidth bottleneck is to operate directly on JPEG-compressed CT volumes, storing 8-bit grayscale data, which retain clinically meaningful diagnostic information at moderate compression ratios (quality factor $\geq 50$, compression ratio $\leq 15{:}1$) \cite{flint2012determining}. Despite the steady growth of telemedicine, however, this direction has remained largely unexplored, since research on chest CT has largely relied on uncompressed, high-fidelity inputs \cite{hamamci2026generalist,pai2025vision,wald2026_colipri,shui2025large,wu2024voco}. Compressed CT therefore offers a promising route to closing the resource-efficiency gap, yet the irreversible loss of spatial information it introduces has, until now, discouraged systematic investigation. It is infeasible to pretrain a model from scratch on large compressed CT corpora under limited-computation-resource constraints, therefore, we focus our attention on leveraging the knowledge of a pretrained model capable of understanding uncompressed CT volumes. This scenario motivates the central research question of our work: \textit{To what extent can the inherent capability to understand high-fidelity volumetric medical images of a pretrained model be transferred to another model for interpreting CT volumes degraded by lossy compression?}

We show that naively fine-tuning a high-fidelity foundation model on JPEG-compressed inputs yields markedly degraded representations, because the teacher's learned features are tightly bound to the spectral and structural characteristics of uncompressed volumes. Conventional knowledge distillation strategies \cite{hinton2015dist}, including naive output matching and feature-level alignment, are insufficient when teacher and student receive inputs that are \emph{structurally similar but spectrally different} \cite{tang2024direct}. On top of that, fundamental distillation strategies, specifically designed for CNN-based models, often underperform in case of ViT-based architectures due to the vast difference between the internal state representations of the two model families \cite{yang2024vitkd}.  To address this, we reframe the cross-fidelity distillation problem as one of style transfer. Our proposed \textbf{Feature Attention Style Transfer (FAST)} framework asks the student to reproduce not only the activation patterns of a frozen high-fidelity teacher (CT-ViT from CT-CLIP, belonging to the class of spatiotemporal vision transformer \cite{hamamci2026generalist}), but also the structural relationships encoded within its attention space, enabling robust feature extraction even when the input has been heavily degraded by compression.

Even with an effective student encoder, deploying contrastive multimodal CT models remains computationally expensive, as dense projection heads account for a substantial share of the overall parameter budget. In CT-CLIP, for instance, flattening a $16{\times}16{\times}16{\times}512$ visual feature tensor into a 512-dimensional embedding produces a single linear layer of approximately $1.07$ billion parameters. To reduce this overhead, inspired by the concepts of Block Tensor Train (BTT) \cite{qiu2024compute}, we replace the dense projection with \textbf{Structured Factorized Projection (SFP)}, cutting the projection-head parameter count by almost half while preserving downstream performance. Together, FAST and SFP form \textbf{CT-Lite}, a contrastive learning pipeline that aligns visual and textual representations of compressed chest CT volumes through a sigmoid-based contrastive objective \cite{zhai2023siglip}, chosen for its tolerance to the small batch sizes imposed by limited GPU memory.

In summary, the contributions of this work are as follows:

\begin{itemize}
    \item \textbf{Feature Attention Style Transfer (FAST):} A distillation framework for spatiotemporal vision transformer that combines Gram-matrix-based attention style preservation with dual-attention feature alignment, enabling robust feature extraction from JPEG-compressed CT volumes.
    \item \textbf{Structured Factorized Projection (SFP):} A Block-Tensor-Train-based projection head that reduces dense projection-layer parameters by almost half while preserving downstream accuracy.
    \item \textbf{First systematic study of compressed-CT diagnostic learning:} We demonstrate that CT-Lite recovers most of the diagnostic signal of uncompressed CT analysis under JPEG compression across three datasets (CT-RATE, NIDCH, RAD-ChestCT), and substantially outperforms compressed-input fine-tuning of both CT-CLIP and COLIPRI.
\end{itemize}

\section{Related Works}

The methodology proposed in this work draws on three intersecting strands of prior research. First, the historical trajectory of thoracic imaging AI helps explain why chest CT analysis has only recently caught up with chest radiography, why high-fidelity volumetric inputs are taken for granted across the modern literature, and why a compressed-input pathway is timely. Second, the literature on knowledge distillation, together with its conceptual connection to neural style transfer, provides the foundation for our cross-fidelity feature alignment strategy. Third, recent advances in tensor-decomposition-based model compression motivate the parameter-efficient projection head we adopt in place of conventional dense layers.

\subsection{Trends in Thoracic Radiology and AI}

Early progress in AI for thoracic imaging is dominated by chest radiography, primarily because of the availability of large-scale public X-ray datasets, such as ChestX-ray14 \cite{wang2017chestxray14}, CheXpert \cite{irvin2019chexpert}, MIMIC-CXR \cite{johnson2019mimiccxr}, PadChest \cite{bustos2020padchest}, and VinDr-CXR \cite{nguyen2022vindrcxr}, emerged well before any comparable resource for chest CT. This data asymmetry persisted for years. Although CT volumes carry substantially more diagnostic information than 2D X-Ray images, the absence of large annotated CT datasets restricted the progress of CT-focused AI research. The Rad-ChestCT dataset \cite{draelos2021machine} represented an early step toward closing this gap by providing expert multi-abnormality labels for chest CT, but it included no paired free-text reports, leaving multimodal volume-language modeling out of reach. The release of CT-RATE \cite{hamamci2026generalist}, the first large-scale chest CT volume-report corpus, marks the point at which CT-based AI research began to catch up with its X-ray counterpart.

Architecturally, the field has traversed three broad eras. The first is CNN-dominated: convolutional backbones such as ResNet \cite{he2016resnet} and DenseNet \cite{huang2017densenet} inspired notable studies for chest X-ray classification \cite{rajpurkar2017chexnet,baltruschat2019comparison}. Draelos et al. extended this design to volumetric CT with CT-Net ~\cite{draelos2021machine}. The second era is reshaped by Vision Transformers (ViTs) \cite{dosovitskiy2021an}, which are quickly adopted in medical imaging through DeiT-style supervised training \cite{touvron2021deit,touvron2022deit3,park2022multitaskvit,mondal2022xvitcos} and self-supervised pre-training strategies including masked autoencoding \cite{he2022mae,xiao2023delvingmae,zhou2023selfpretrainmae} and contrastive learning \cite{he2020moco,chen2020simclr,caron2021dino,sowrirajan2021mococxr,azizi2021bigssl,azizi2023remedis}. Spatiotemporal extensions of ViTs \cite{arnab2021vivit,bertasius2021timesformer,fan2021mvit} provided the architectural foundation for processing volumetric data, and underpin the CT-ViT encoder used as the high-fidelity teacher in the present work \cite{hamamci2026generalist}.

The third and current era is defined by multimodal and vision--language pre-training. Following CLIP \cite{radford2021clip} and SigLIP \cite{zhai2023siglip}, report-supervised contrastive learning enabled scalable representation learning in chest radiography---ConVIRT \cite{zhang2020convirt}, GLoRIA \cite{huang2021gloria}, BioViL \cite{boecking2022biovil}, MedCLIP \cite{wang2022medclip}, and zero-shot extensions such as CheXzero \cite{tiu2022chexzero}. The same paradigm has now been extended to chest CT, beginning with CT-CLIP \cite{hamamci2026generalist} and followed by CT-FM \cite{pai2025vision}, COLIPRI \cite{wald2026_colipri}, fVLM \cite{shui2025large}, scalable 3D vision--language pre-training \cite{zhao2025towards}, and volume contrastive learning (VoCo) \cite{wu2024voco}. Knowledge transfer from 2D X-ray experts to 3D CT models---BIUD \cite{cao2024bootstrapping} and ViSD-Boost \cite{cao2025boosting}---and multimodal large language models for radiology \cite{nam2025multimodal} \cite{Blankemeieretal2026} reflect ongoing efforts to bridge the residual data gap and broaden clinical applicability. Comprehensive surveys situate this rapidly growing literature \cite{hayat2022vlpsurvey,tang2024multimodal}.

A common assumption underlying every model in this lineage is that input volumes are stored in their original uncompressed form, which keeps the data pipeline computationally heavy and bandwidth-intensive. CT-Lite extends this lineage into the compressed-input regime, using CT-CLIP's CT-ViT as the high-fidelity teacher. We adopt CT-CLIP rather than the more recent COLIPRI \cite{wald2026_colipri}, or fVLM \cite{shui2025large} because its weights are the only publicly available 3D chest CT representation at the time of our study; our primary objective is not to surpass the latest foundation model but to demonstrate that a student trained on compressed inputs can closely track the established high-fidelity teacher under tight resource constraints.

\subsection{Knowledge Distillation (KD) and Neural Style Transfer}

Knowledge distillation (KD), first formalized by Hinton et al.~\cite{hinton2015dist}, enables the transfer of learned representations from a teacher model to a student model, typically yielding compact and efficient networks. Beyond the original logit-matching formulation, feature-level distillation \cite{romero2015fitnets}, attention transfer \cite{zagoruyko2017attentiontransfer}, contrastive representation distillation \cite{tian2019crd} and relational distillation \cite{park2019relational} have broadened the scope of knowledge transfer. 

In the context of transformers, DeiT \cite{touvron2021deit,touvron2022deit3} demonstrated that distillation tokens can effectively transfer knowledge within ViT architectures, and subsequent work has refined ViT-specific distillation through feature-level alignment \cite{yang2024vitkd}, weight-multiplexing for compact students \cite{zhang2022minivit}, scalable strong-teacher schemes \cite{fan2024scalekd}, and self-distillation with restructured teacher--student configurations \cite{zheng2024restructuring}. However, effective KD for \emph{spatiotemporal} transformers \cite{arnab2021vivit,bertasius2021timesformer,fan2021mvit} critical for volumetric medical data remains comparatively underexplored. In the medical domain, heterogeneous KD using information flow modeling \cite{passalis2020heterogeneous} enables transfer across architectures, while UniCompress \cite{yang2024unicompress} applies distillation to improve medical image compression without sacrificing diagnostic fidelity. Usmani et al.~proposed STKD-VViT for video deepfake detection \cite{usmani2025spatio}, a methodology with parallels for dynamic medical imaging.

A further departure from convention is that nearly all of the above frameworks assume the canonical KD setup: a heavyweight teacher and a lightweight student that consume the \emph{same} input. CT-Lite inverts this convention along the input axis rather than the capacity axis---teacher and student share an \emph{identical} architecture but receive \emph{different} inputs (uncompressed and compressed CT volumes), so the distillation signal must bridge a fidelity gap rather than a capacity gap.

A second line of work that directly informs our methodology is neural style transfer, introduced by Gatys et al.~\cite{gatys2016image}, which showed that the Gram matrix of CNN feature activations captures the second-order statistics of an image's ``style'' independently of its content. The Gram-matrix construction has since been used wherever the goal is to align the structural co-activation patterns of two feature distributions rather than their pointwise values. The attention-style preservation loss in our Feature Attention Style Transfer (FAST) framework is a direct adaptation of this idea: when teacher and student receive inputs that are structurally similar but spectrally different, attention maps cannot be matched value-for-value, but the relational pattern they encode, captured by their Gram matrix, can. We thus treat the teacher's attention patterns as the ``style'' to be preserved across changes in input fidelity and combine this signal with dual-attention feature alignment to transfer both activation patterns and structural relationships from high-fidelity to compressed CT representations.

\subsection{Model Compression and Efficient Architectures}

Deploying deep learning models in resource-constrained clinical environments necessitates techniques for reducing parameter count and inference cost. Magnitude and connection pruning \cite{han2015learning,han2016deepcompression} and the lottery-ticket hypothesis \cite{frankle2019lottery} demonstrated that large networks can be aggressively sparsified with little accuracy loss, while integer-arithmetic quantization \cite{jacob2018quantization} and post-training/quantization-aware variants for large transformers \cite{frantar2023gptq,dettmers2023qlora} extend this efficiency to inference- and finetuning-time. Low-rank weight factorization, typically realized via singular value decomposition, offers a complementary axis of compression \cite{sainath2013lowrank,denton2014exploiting}, and its parameter-efficient adaptation form, Low-Rank Adaptation (LoRA) \cite{hu2022lora}, has gained widespread adoption for fine-tuning of large language models by injecting trainable low-rank matrices. Knowledge distillation \cite{gou2021kdsurvey} provides a further complementary route by transferring representations from larger teachers to smaller students.

While attention-matrix compression has received considerable focus, dense (fully connected) layers often constitute the dominant parameter sink. Tensor decomposition methods---including Kronecker products \cite{sindhwani2015structured,edalati2022krona}, Tucker decomposition \cite{kim2016tucker}, and Tensor Train (TT) \cite{oseledets2011tt,novikov2015tensorizing}---offer principled approaches to compactly parameterizing large weight matrices. Among these, Block Tensor Train (BTT) decomposition \cite{qiu2024compute} has been shown to be particularly effective, enabling depth expansion while maintaining stable computational costs.

Our Structured Factorized Projection (SFP) builds on this line of work to address the parameter overhead of projection heads in contrastive learning. SFP replaces the  linear projection layer in contrastive learning with a sum of BTT-factorized blocks, reducing the projection-head parameter count by almost half while preserving downstream performance.

\section{Methodology}

\subsection{Stage 1: Feature-Attention Style Transfer (FAST)}
Let, $\Phi_{vis,T}$ and $\Phi_{vis,S}$ denote the teacher and student visual encoder, respectively. The teacher encoder $\Phi_{vis,T}$ is pretrained to extract rich visual features given a set of uncompressed high-fidelity chest CT volumes, \{$\mathcal{X}_U$\}. The student encoder $\Phi_{vis,S}$ receives degraded input set \{$\mathcal{X}_C$\}, obtained from applying lung window (window center = -600 HU and window width=1500 HU) to each sample in \{$\mathcal{X}_U$\}, and subsequently compressing to JPEG. The objective is to train $\Phi_{vis,S}$ to enable it extract such visual features from degraded inputs {$\mathcal{X}_C$} that is similar or very close to what $\Phi_{vis,T}$ extracts from the high-fidelity inputs \{$\mathcal{X}_U$\}, as given by equation \ref{objective}.

\begin{figure*}[!t]
    \centering
    \includegraphics[width=\textwidth]{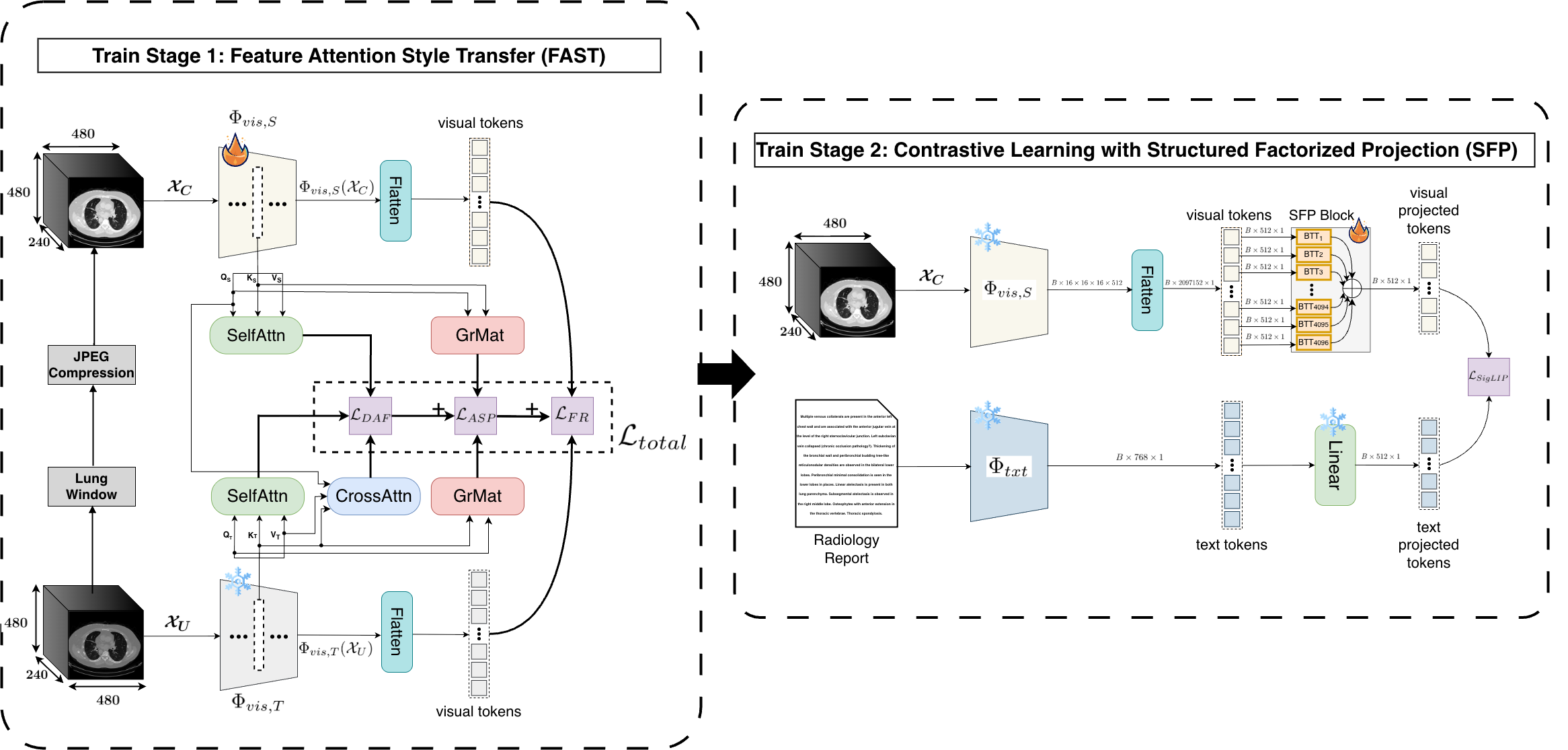}
    \caption{Overview of the CT-Lite framework. \textbf{(Left)} Stage 1: Feature Attention Style Transfer (FAST) distills the frozen high-fidelity teacher into a student encoder operating on JPEG-compressed inputs, using attention-style and dual-attention feature alignment. \textbf{(Right)} Stage 2: contrastive alignment of compressed-CT and report embeddings via Structured Factorized Projection (SFP) blocks and a SigLIP objective.}
    \label{fig:methodology_overview}
\end{figure*}

\begin{equation}\label{objective}
    \bm{\theta}_S^* = \arg\min_{\bm{\theta}_\student}  MSE(\Phi_{vis,T}({\mathcal{X}}_U; {\theta}_T), \Phi_{vis,S}({\mathcal{X}}_C; {\theta}_S)) 
\end{equation}

Here, $\theta_T$, $\theta_S$, $\theta_S^*$ and $MSE$ represent teacher parameters, student parameters, optimal student parameters and Mean Squared Error (MSE), respectively.

We isolate the visual encoder (CT-ViT) from CT-CLIP and utilize it as our choice of both the teacher and the student encoder, where a series of spatial and causal transformer blocks enable attention calculation across all three dimensions (height, width, depth). The teacher encoder is initialized with the pretrained weights and kept frozen during training, while the student encoder is randomly initialized. The training loss that satisfies the objective in Equation \ref{objective} depends on how the information is processed in each layer of CT-ViT. 

To determine the optimal loss function at first, five independent experiments are carried out for 50 epochs with a small dataset of 50 CT volumes (See Figure~\ref{fig:ablation_convergence}). AdamW optimizer with cosine annealing scheduler with a learning rate of $1 \times 10^{-5}$ is used across all five experiments. The first approach is the naive approach that tries to directly align the end representations between the teacher and the student by calculating MSE between them, in which the student model seemed to converge with a high degree of error. A higher degree of mismatch is observed between the intermediate representations of the student and the teacher in the first experiment, which led us to apply feature-level distillation as our second approach, where the student tries to mimic the internal feature representations of the teacher model. The loss function in the first approach also acted as a component in the second approach. This approach is faced with a slower rate of convergence, forcing us to rule it out as infeasible for a larger dataset under low-resource constraint.

We hypothesize that the failure of conventional distillation approaches arises from the fact that the student and the teacher, in our case, receive essentially different inputs that may have different voxel values but retain a higher degree of structural similarity. This led us to find the similarity of the problem to a neural style transfer problem. We combined elements of feature-level distillation and neural style transfer to propose \textbf{Feature-Attention Style Transfer (FAST)} framework, which enforces the student to mimic not only the activation patterns but also the structural relationships encoded within the teacher's attention space.  The \textit{FAST} approach results in a much faster alignment between the encoded outputs of the student and the teacher.

\textit{FAST} incorporates 3 loss components. The rest of this subsection discusses the components in detail. 

\subsubsection{Attention Style Preservation Loss ($\mathcal{L}_{ASP}$)}
This component focuses on the relational structure of the attention maps rather than their direct values. We employ a Gram-matrix-based loss to distil the structural 'style' of the attention maps. Given attention score $\bm{A_{score}}= softmax(\bm{QK}^T/\sqrt{d_k})$, $\mathcal{L}_{ASP}$ is calculated by:

\begin{equation}
    \mathcal{L}_{ASP} = ||\bm{G}_T - \bm{G}_S||
\end{equation}

where $\bm{G}$ = $\bm{A}_{score}$$\bm{A}_{score}^T$

This loss captures the pairwise correlations between all query distributions.

\subsubsection{Dual-Attention Feature Loss ($\mathcal{L}_{DAF}$)}
This loss component enforces alignment between the output of the attention mechanisms, given by:

\begin{equation}
\begin{aligned}
\mathcal{L}_{DAF}
  &= \mathrm{MSE}\!\left(\mathrm{SA}(\bm{Q}_T,\bm{K}_T,\bm{V}_T),\ \mathrm{SA}(\bm{Q}_S,\bm{K}_S,\bm{V}_S)\right) \\
  &\quad + \mathrm{MSE}\!\left(\mathrm{SA}(\bm{Q}_T,\bm{K}_T,\bm{V}_T),\ \mathrm{CA}(\bm{Q}_S,\bm{K}_T,\bm{V}_T)\right),
\end{aligned}
\end{equation}
where $\mathrm{SA}(\cdot)$ and $\mathrm{CA}(\cdot)$ denote self and cross-attention, respectively.

The first term enforces self-attention alignment, encouraging the student to reproduce the same attention pattern as the teacher. The second term introduces cross-attention fusion, enabling the student to interpret the teacher's attention space. Together, they promote richer feature transfer between the encoders.

\subsubsection{Final Representation Loss}
This attempts to align the end representation by minimizing the naive distillation loss, given by:

\begin{equation}
 \mathcal{L}_{FR} = MSE(\Phi_{vis,T}(\mathcal{X}_U), \Phi_{vis, S}(\mathcal{X}_C))
\end{equation}

 The combined training loss is the linear combination of all three training objectives discussed above, given by:

\begin{equation}\label{eq:FinalLoss}
 \mathcal{L}_{total} = \alpha \mathcal{L}_{ASP} + \beta \mathcal{L}_{DAF} + \gamma \mathcal{L}_{FR}
\end{equation}

It is important to note that both $\mathcal{L}_{DAF}$ and $\mathcal{L}_{ASP}$ are calculated and cumulatively added across all the spatial and temporal transformer blocks in CT-ViT.

\subsection{Stage 2: Contrastive Learning with Structured Factorized Projections (SFP)}
In the original CT-CLIP framework, the encoded outputs of the visual encoder (CT-ViT) and the text encoder (CXR-BERT) \cite{boecking2022making} are flattened and projected into fully connected projection layers to map the encoded outputs into a shared low-dimensional (512) latent space for contrastive image-text alignment. The text projection layer is relatively lightweight since CXR-Bert produces a 768-dimensional flattened embedding and is therefore reasonable in terms of compute and parameter count.


However, the the visual projection head poses a major bottleneck in terms of parameters, compute and memory overhead.

This challenge motivates us to introduce \textbf{Structured Factorized Projection (SFP)} block as a suitable replacement for parameter-heavy dense layers. SFP utilizes the concept of Block Tensor Train (BTT) decomposition.  

In a fully-connected layer, assuming no bias term for simplicity, the Matrix-Vector Multiplication (MVM) operation is given by-

\begin{equation}\label{DenseLayer}
\mathbf{Y = WX}
\end{equation}

Given that input $\mathbf{X} \in \mathbf{R}^{in\_dim \times 1}$ , $\mathbf{W} \in \mathbb{R}^{out\_dim \times in\_dim}$ and $\mathbf{Y} \in \mathbb{R}^{out\_dim \times 1}$. For a special case where $in\_dim = M \times out\_dim$, where $M$ is a positive integer, Equation \ref{DenseLayer} can be rewritten as-

\begin{equation}\label{DenseLayer2}
\mathbf{Y} = \big[\mathbf{W}_1\ \cdots\ \mathbf{W}_M\big]\big[\mathbf{X}_1\ \cdots\ \mathbf{X}_M\big]^{T},
\end{equation}
or equivalently,
\begin{equation}\label{denselayer3}
\mathbf{Y} = \sum_{i=1}^{M} \mathbf{W}_i\,\mathbf{X}_i^{T}.
\end{equation}

where $\mathbf{W}_i \in \mathbb{R}^{out\_dim \times out\_dim}$, $\mathbf{X}_i \in \mathbb{R}^{out\_dim \times 1}$.
We replace each square-dimensional matrix, $\mathbf{W}_i$ with Block-Tensor Train (BTT) matrix $\mathbf{BTT}_i$, so that Equation \ref{denselayer3} becomes-

\begin{equation}\label{denselayer4}
\mathbf{Y} = \mathbf{Y}_{vis} =  \sum_{i=1}^M{\mathbf{BTT}_i \; \mathbf{X}_i^T}  
\end{equation}

The matrix--vector multiplication (MVM) for each block
is computed using a BTT-style structured factorization.
Let $out\_dim = d_1 \times d_2$ be a factorization of the output
dimension, where $d_1, d_2 \in \mathbb{N}$.
Each input chunk $\mathbf{X}_i \in \mathbb{R}^{out\_dim} \times 1$ is reshaped
into a matrix $\mathbf{X}_i \in \mathbb{R}^{d_1 \times d_2}$.

Each structured projection $\mathbf{BTT}_i$ is parameterized by two
core tensors,
$\mathcal{\bm{L}}_i \in \mathbb{R}^{d_1 \times d_2 \times d_1 \times r}$ and
$\mathcal{\bm{R}}_i \in \mathbb{R}^{r \times d_1 \times d_2}$,
where $r$ denotes the BTT rank.
The output matrix $\mathbf{Y}_i \in \mathbb{R}^{d_1 \times d_2}$ is
computed element-wise as

\begin{equation}
\label{eq:btt_mvm}
(\mathbf{Y}_i)_{p,j}
=
\sum_{q=1}^{d_1}
\sum_{k=1}^{d_2}
\sum_{a=1}^{r}
(\mathcal{\bm{L}}_i)_{p,j,q,a}
\,
(\mathcal{\bm{R}}_i)_{a,q,k}
\,
(\mathbf{X}_i)_{q,k},
\end{equation}

where $p,q \in \{1,\dots,d_1\}$, $j,k \in \{1,\dots,d_2\}$, and $r$ is
the BTT rank.
The output matrix $\mathbf{Y}_i$ is then flattened to obtain a vector
in $\mathbb{R}^{out\_dim \times 1}$.

\textbf{Parameter Complexity.} The number of parameters in an SFP layer is given by:
\begin{equation} \label{eq:sfp_params}
\text{Params}_{\text{SFP}} = 2 \times r \times \text{in\_dim} \times \sqrt{\text{out\_dim}}
\end{equation}
where $r$ is the BTT-rank, $\text{in\_dim}$ is the input dimension, and $\text{out\_dim}$ is the output dimension.

\textbf{Efficiency Bound.} For SFP to yield a parameter reduction over a standard dense layer, the condition $2 r \times \text{in\_dim} \times \sqrt{\text{out\_dim}} \leq \text{in\_dim} \times \text{out\_dim}$ must hold, which simplifies to
\begin{equation} \label{eq:rank_bound}
r \leq \frac{\sqrt{\text{out\_dim}}}{2}.
\end{equation}
For our visual projection head with $\text{out\_dim} = 512$, this implies $r \leq \sqrt{512}/2 \approx 11.3$. Our resources allow us to experiment with SFP block with a maximum rank of 6. For $\text{in\_dim} = 16 \times 16 \times 16 \times 512 = 2{,}097{,}152$ and $r = 6$, SFP yields approximately $2 \times 6 \times 2{,}097{,}152 \times \sqrt{512} \approx 569$M FLOPs (and parameters), , almost half of CT-CLIP's dense projection head (1.07B). 

\textbf{Weight Initialization via SVD-based Approximation.}
While the SFP layers can be trained from scratch, they can also be initialized by approximating a pre-trained dense weight matrix $\bm{W} \in \mathbb{R}^{d_{out} \times d_{in}}$ from a teacher model like CT-CLIP \cite{hamamci2026generalist}. To initialize the core tensors $\{\mathcal{\bm{L}}_i, \mathcal{\bm{R}}_i\}_{i=1}^M$, we first partition the dense weights into $M$ sub-matrices $\bm{W}_i \in \mathbb{R}^{d_{out} \times d_{out}}$. For each block, we perform a truncated Singular Value Decomposition (SVD) to find a low-rank approximation of rank $r$:
\begin{equation}
\bm{W}_i \approx \bm{U}_i \bm{\Sigma}_i \bm{V}_i^\top \implies \bm{W}_i \approx \hat{\bm{U}}_i \hat{\bm{V}}_i^T,
\end{equation}
where $\hat{\bm{U}}_i = \bm{\bm{U}}_{i, :r} \sqrt{\bm{\Sigma}_{i, :r}}$ and $\hat{\bm{V}}_i^\top = \sqrt{\bm{\Sigma}_{i, :r}} \bm{V}_{i, :r}^T$. The core tensors are then initialized such that their contraction aligns with this low-rank structure:
\begin{equation}
(\bm{\mathcal{L}}_i)_{p,j,q,a} \gets \text{reshape}(\hat{\bm{U}}_i), \quad (\bm{\mathcal{R}}_i)_{a,q,k} \gets \text{reshape}(\hat{\bm{V}}_i^T).
\end{equation}
This initialization ensures that the model starts with a representation space that is functionally close to CT-CLIP's high-fidelity projection \cite{hamamci2026generalist}.

\begin{algorithm}
\caption{Structured Factorized Projection (SFP)}
\label{alg:sfp}
\begin{algorithmic}[1]
\Require Input feature vector $\mathbf{X} \in \mathbb{R}^{in\_dim \times 1}$
\Require Number of blocks $M$, output dimension $out\_dim$
\Require Output factorization $out\_dim = d_1 \times d_2$
\Require BTT-style core tensors $\{(\bm{\mathcal{L}}_i, \bm{\mathcal{R}}_i)\}_{i=1}^{M}$
\Ensure Projected output vector $\mathbf{Y} \in \mathbb{R}^{out\_dim \times 1}$

\State Partition the input vector:
\[
\mathbf{X} =
\begin{bmatrix}
\mathbf{X}_1 \\
\mathbf{X}_2 \\
\vdots \\
\mathbf{X}_M
\end{bmatrix},
\qquad
\mathbf{X}_i \in \mathbb{R}^{out\_dim \times 1}
\]

\State Initialize output vector:
\[
\mathbf{Y} \leftarrow \mathbf{0} \in \mathbb{R}^{out\_dim \times 1}
\]

\For{$i = 1$ to $M$}
    \State Reshape input block:
    \[
    \mathbf{X}_i \leftarrow \mathrm{reshape}(\mathbf{X}_i, d_1, d_2), \:  so \: \ \mathbf{X}_i \in \mathbb{R}^{out\_dim \times 1}
    \]
    
    \State Compute BTT block output using Eq.~\eqref{eq:btt_mvm}:
    \Statex \centering
    $\begin{aligned}
    (\mathbf{Y}_i)_{p,j}
    &=
    \sum_{q=1}^{d_1}
    \sum_{k=1}^{d_2}
    \sum_{a=1}^{r}
    (\bm{\mathcal{L}}_i)_{p,j,q,a}
    (\bm{\mathcal{R}}_i)_{a,q,k}
    (\mathbf{X}_i)_{q,k}
    \end{aligned}$

    \State \raggedright Vectorize block output:
    \[
    \mathbf{Y}_i \leftarrow \mathrm{vec}(\mathbf{Y}_i) \in \mathbb{R}^{out\_dim \times 1}
    \]
    
    \State Accumulate output:
    \[
    \mathbf{Y}_{vis} \leftarrow \mathbf{Y} + \mathbf{Y}_i
    \]
\EndFor

\State \Return $\mathbf{Y}$
\end{algorithmic}
\end{algorithm}

The projected visual output is then aligned with the corresponding text projection vectors using SigLIP loss \cite{zhai2023siglip}, which treats each image--text pair within a mini-batch as an independent binary classification problem. Unlike the softmax-based CLIP objective, SigLIP does not rely on competition among in-batch negatives and is therefore less sensitive to batch size, making
it well suited for training under limited GPU memory.

\begin{figure}[pos=h]
    \includegraphics[width=\columnwidth]{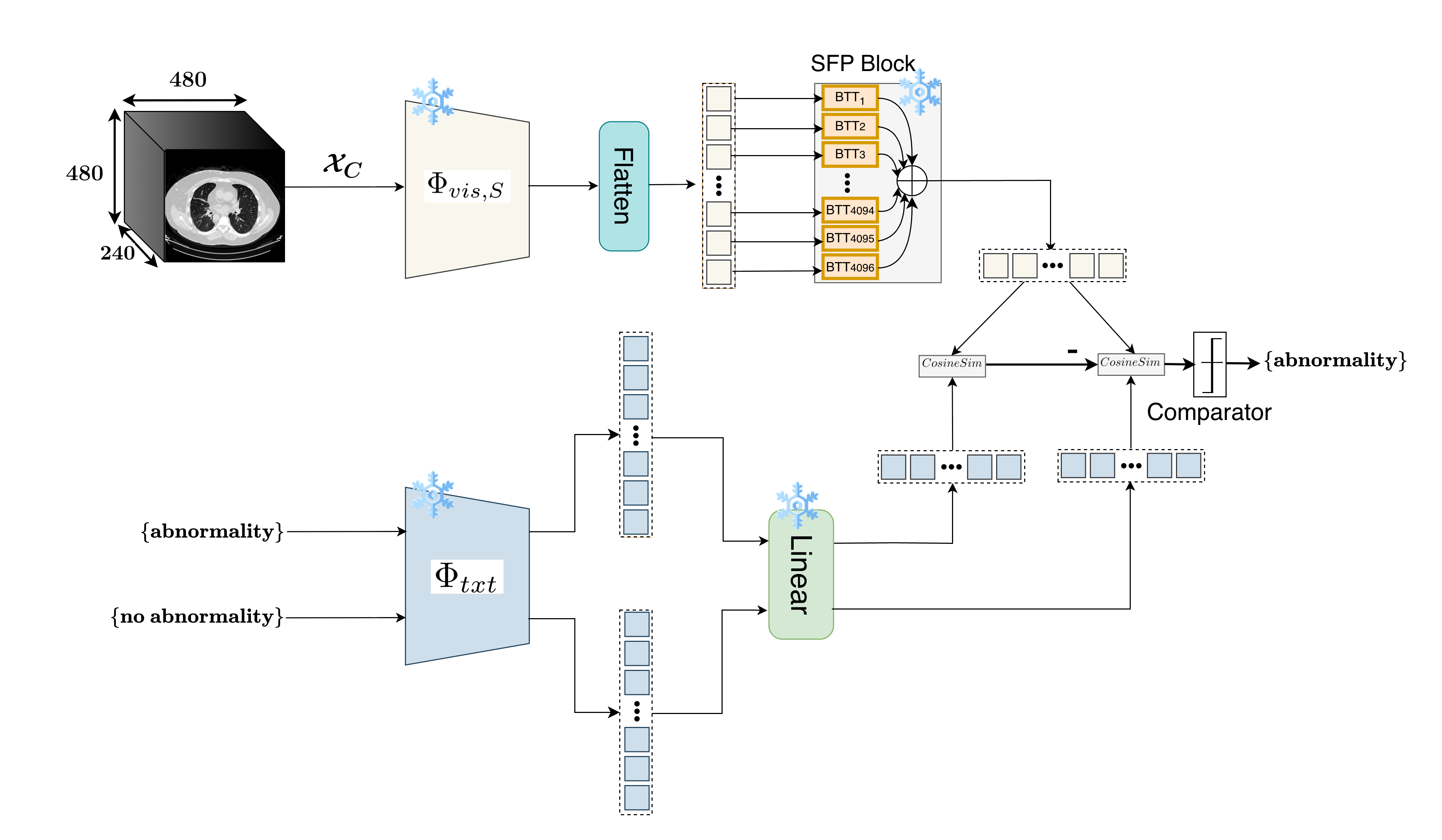}
    \caption{Inference strategy of CT-Lite using contrasting prompts}
    \label{fig:inference_strategy}
\end{figure}

\subsection{Inference Strategy}
Contrasting textual prompts are employed to detect each abnormality using CT-Lite, given an input CT volume (See Figure \ref{fig:inference_strategy}). The visual encoder of CT-Lite, combined with the SFP block, generates a visual embedding from the input, while the text encoder and latent projection layer generate text embeddings for a positive prompt and a negative prompt (e.g., "emphysema" and "no emphysema"). The cosine similarity between the visual embedding and each text embedding is computed, and the abnormality is predicted as present if the difference between the two similarity scores exceeds a given threshold. The threshold is set using Youden's J statistic.


\section{Experiments and Results}

\subsection{Datasets}

\textbf{CT-RATE:} CT-RATE is the first publicly available large-scale dataset comprising chest CT volumes paired with their corresponding radiology reports, collected from Istanbul Medipol University Mega Hospital. We sampled a subset of 1,837 CT volume–report pairs, corresponding to 788 unique patients, and incorporated them into our study. The sampled data are split into non-overlapping training and validation sets, consisting of 1,333 volumes from 573 patients and 504 volumes from 215 patients, respectively. Each volume in the dataset is provided in uncompressed NIfTI format, accompanied by metadata attributes such as voxel spacing, rescale slope, and rescale intercept. The CT scans are acquired using scanners manufactured by Philips and Siemens Healthineers. \\

\textbf{NIDCH Chest CT Dataset:} In order to introduce robustness and generalizability, we retrospectively collected and curated a dataset of 1847 JPEG-compressed (Q-factor = 90\%) stacks of CT slices paired with their radiology reports, corresponding to 681 unique patients from National Institute of Diseases of the Chest and Hospital (NIDCH), Bangladesh from December 2023 to July 2024. The CT scanner model used in this hospital is a 3rd Generation Hitachi Scenaria 128-Slice CT scanner, and the workstation uses Fujifilm Synapse 3D analyzer. The training and validation splits contain 1336 and 511 volumes, corresponding to 505 and 176 unique patients, respectively. The corresponding high-fidelity counterparts of all the volumes in the validation split are also collected (in anonymized DICOM file format) for conducting comparison against the baseline.\\

\textbf{RAD-ChestCT:} RAD-ChestCT \cite{draelos2021machine} is a publicly available dataset designed for radiology report generation and multi-label abnormality detection in chest CT scans. In addition to the two data sources mentioned before, we sample 501 instances from the RAD-ChestCT dataset as an external validation dataset to compare against benchmarks. The CT volumes are available in uncompressed numpy zip (.npz) format.

\subsection{Implementation Details}
CT-Lite is trained on an NVIDIA L4 GPU, while the rest of the fine-tuning for evaluation is carried out on an NVIDIA A100 GPU; both accessed through Google Colab Pro+.

\begin{table}[pos=h]
\centering
\caption{Training Hyperparameters for FAST and SFP Stages}
\label{tab:hyperparameters}
\begin{tabular}{lc}
\toprule
\textbf{Hyperparameter} & \textbf{Value} \\
\midrule
\multicolumn{2}{c}{\textbf{Stage 1: Feature Attention Style Transfer (FAST)}} \\
\midrule
Optimizer                              & AdamW \\
Weight Decay                           & $0.01$ \\
Learning Rate                          & $5 \times 10^{-5}$ \\
Scheduler                              & Cosine Annealing \\
Minimum Learning Rate ($\eta_{min}$)   & $1 \times 10^{-6}$ \\
Batch Size                             & $4$ \\
Epochs                                 & $20$ \\
Loss Weights ($\alpha, \beta, \gamma$) & $2,\ 5,\ 22$ \\
\midrule
\multicolumn{2}{c}{\textbf{Stage 2: Contrastive Learning with SFP}} \\
\midrule
Block Tensor Train (BTT) Rank ($r$)    & $6$ \\
Number of Blocks ($M$)                 & $4096$ \\
Optimizer                              & AdamW \\
Weight Decay                           & $0.01$ \\
Base Learning Rate                     & $1 \times 10^{-5}$ \\
Effective Learning Rate ($\eta_{SFP}$) & $8.53 \times 10^{-4}$ \\
Scheduler                              & Cosine Annealing \\
Minimum Learning Rate ($\eta_{min}$)   & $1 \times 10^{-6}$ \\
Batch Size                             & $4$ \\
Epochs                                 & $10$ \\
Objective                              & SigLIP \\
\bottomrule
\end{tabular}
\end{table}

\textbf{Hyperparameter Scaling:}
To maintain training stability when transitioning from the dense projection head ($in\_dim = 2{,}097{,}152$) to the SFP block, we apply the principles of Maximal Update Parameterization ($\mu$P) \cite{yang2021tuning}. The learning rate $\eta_{SFP}$ is chosen to ensure that the update magnitude remains invariant to the change in effective width. Given the original dense learning rate $\eta_{base}$, the SFP learning rate is scaled by the ratio of the fan-ins:
\begin{equation}
    \eta_{SFP} = \eta_{base} \times \frac{d_{in}}{M \times r}
\end{equation}
For $r=6$ and $M=4096$, the effective fan-in is reduced from $\sim 2.1 \times 10^6$ to $24{,}576$. This necessitates a scaling factor of approximately $85.33\times$ over the base dense learning rate to compensate for the structural bottleneck and maintain consistent update magnitudes across parameterizations.

The training process is divided into two distinct stages: Feature Attention Style Transfer (FAST) and Structured Factorized Projection (SFP) contrastive alignment. The detailed hyperparameters for both stages are summarized in Table~\ref{tab:hyperparameters}. Both stages utilize the AdamW optimizer with a weight decay of $0.01$ and a cosine annealing learning rate scheduler ($\eta_{min} = 10^{-6}$) to ensure stable convergence. Due to the high memory demands of processing volumetric data, the batch size is restricted to 4 for both FAST and SFP stages, respectively.

For the FAST stage, the loss weighting coefficients, after observing the individual loss component magnitudes for a number of inputs, are carefully set to $\alpha = 2$, $\beta = 5$, and $\gamma = 22$, so that the loss components in Equation \ref{eq:FinalLoss} maintain a tentative ratio of 1:1:4. 

For the SFP stage, establishing an appropriate learning rate is critical when transitioning from a dense projection head to a factorized architecture. Following the $\mu$P principles described above, the original dense layer possessed a fan-in of $2{,}097{,}152$. By setting the BTT rank to $r=6$ across $M=4096$ blocks, the effective SFP fan-in is reduced to $24{,}576$. To compensate for this structural bottleneck, the base learning rate of $1 \times 10^{-5}$ is multiplied by the fan-in ratio ($\approx 85.33$), yielding an adjusted SFP learning rate of $8.53 \times 10^{-4}$.

\begin{figure*}[!t]
    \centering

    \begin{subfigure}{0.48\textwidth}
        \centering
        \includegraphics[width=\linewidth]{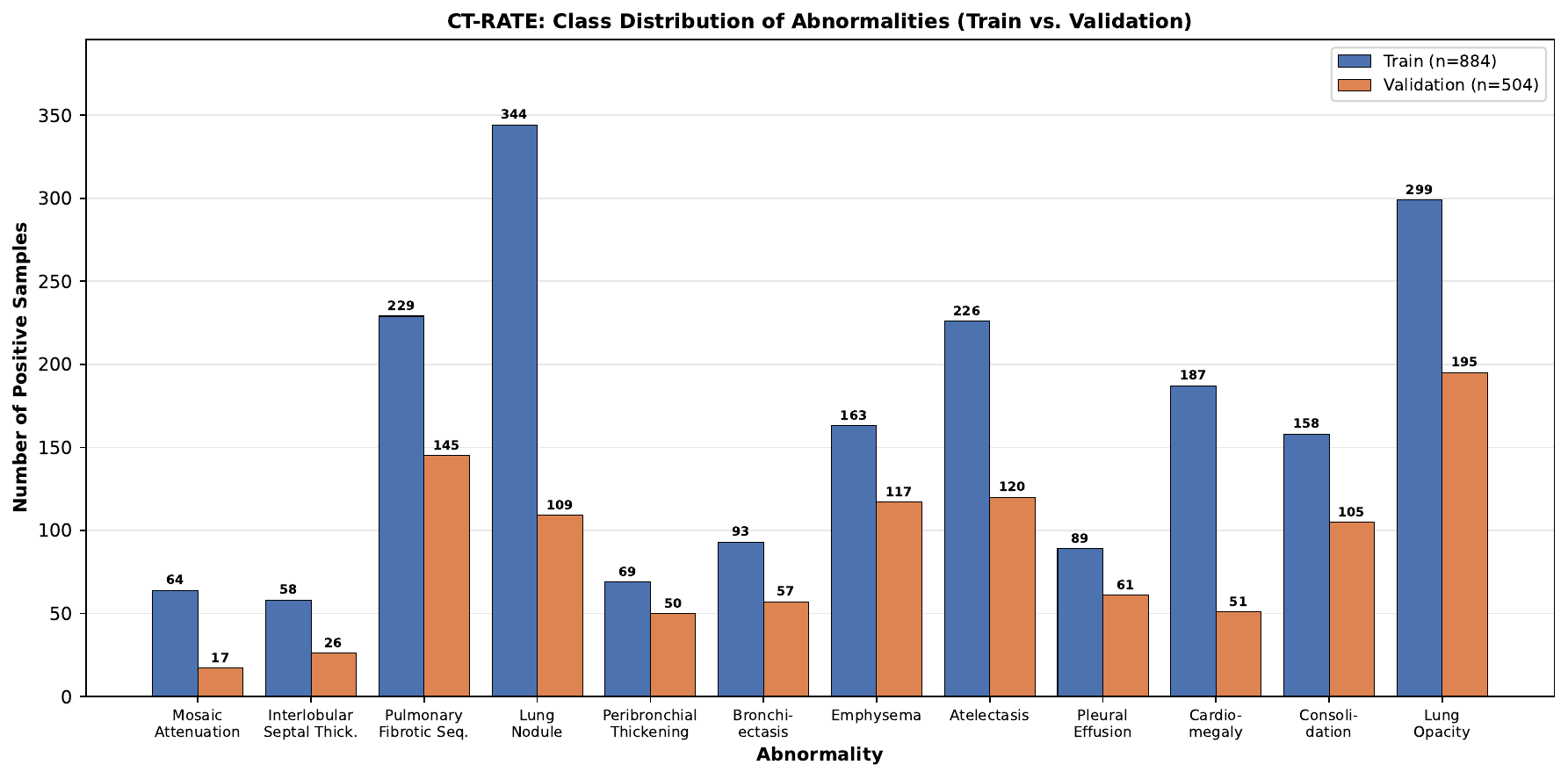}
        \caption{CT-RATE (subset used for training and validation)}
        \label{fig:hist_ctrate}
    \end{subfigure}
    \hfill
    \begin{subfigure}{0.48\textwidth}
        \centering
        \includegraphics[width=\linewidth]{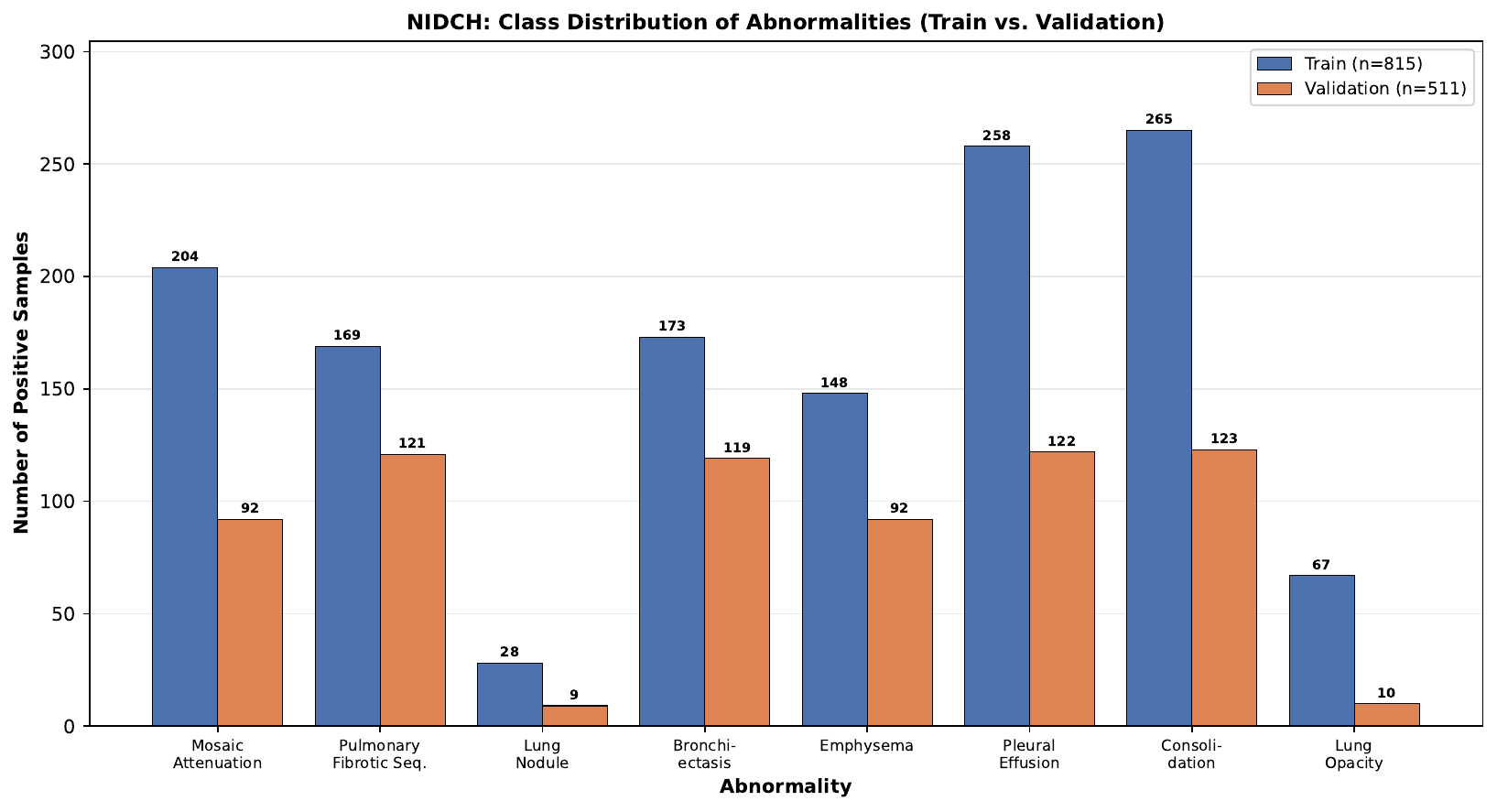}
        \caption{NIDCH}
        \label{fig:hist_nidch}
    \end{subfigure}

    \vspace{0.4cm}

    \begin{subfigure}{0.48\textwidth}
        \centering
        \includegraphics[width=\linewidth]{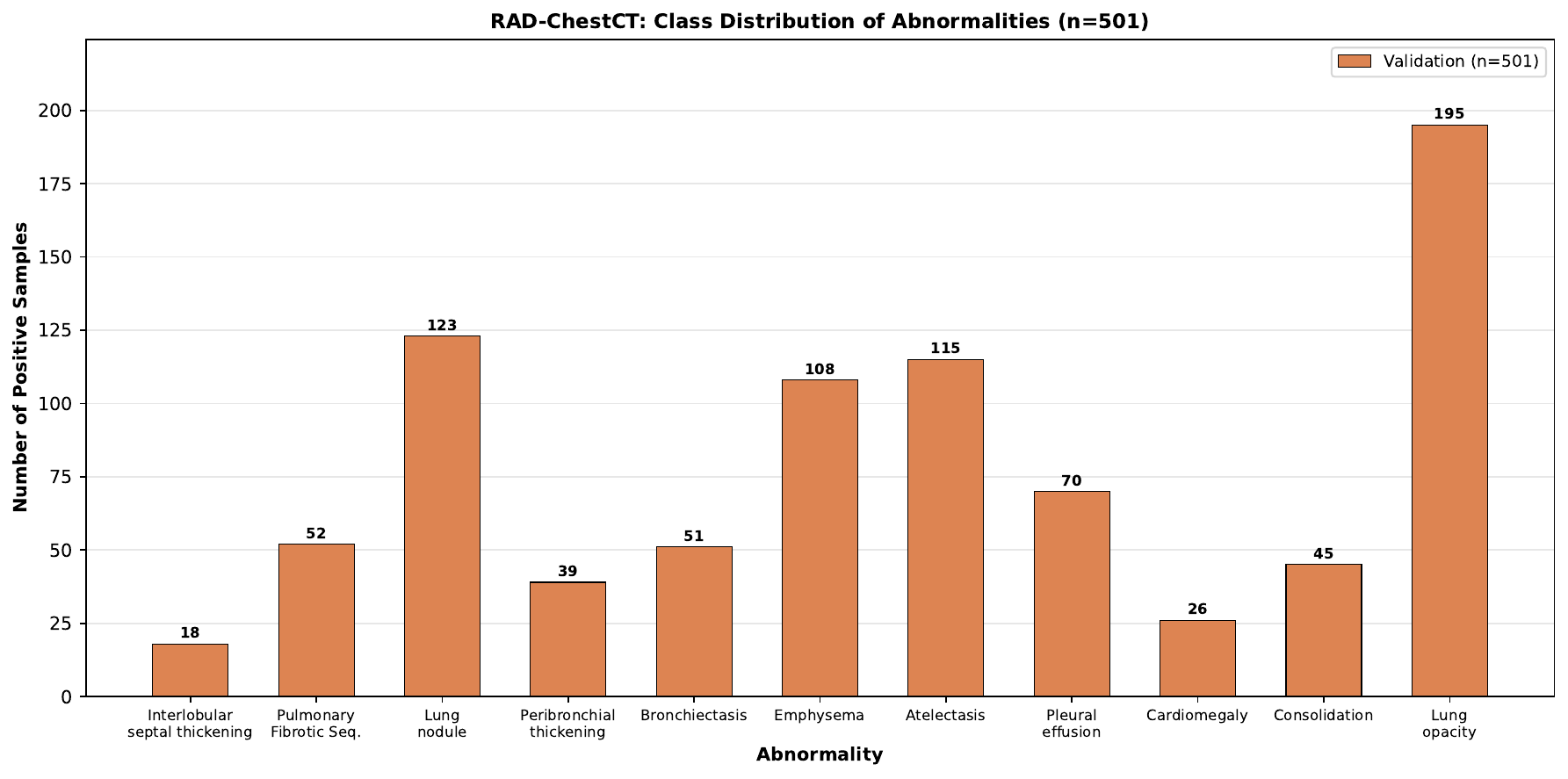}
        \caption{RAD-ChestCT (subset used for validation)}
        \label{fig:hist_radchestct}
    \end{subfigure}

    \caption{Class distribution histograms for the three evaluation datasets. CT-RATE and RAD-ChestCT distributions correspond to the subsets employed in this study, not the complete original datasets.}
    \label{fig:class_histograms}
\end{figure*}

Figure~\ref{fig:class_histograms} illustrates the class distribution across the three datasets. 

\subsection{Image Preprocessing}

The FAST training stage requires both uncompressed high-fidelity CT volumes as well as their degraded JPEG-compressed counterparts. Like CT-CLIP, the CT volumes are resized to achieve an anisotropic voxel spacing of 0.75 mm $\times$ 0.75 mm $\times$ 1.5 mm. Each of the CT volumes is subsequently converted to Hounsfield Unit (HU) values using the rescale slope and rescale intercept from the metadata attributes. The voxel-wise transformation is given by-

\begin{equation}
\label{eq:huconv}
    P_{\mathrm{HU}}(x,y,z) = m \cdot P(x,y,z) + b,
\end{equation}
where $P(x,y,z)$ denotes the voxel value at coordinates $(x,y,z)$, and $m$ and $b$ are the rescale slope and rescale intercept obtained from the volume metadata. For feeding the teacher encoder of the FAST stage, the volumes are clipped to HU values [-1000, 1000], followed by normalization to [-1,1]. All compressed data used in our pipeline are obtained by applying lung-windowing to the original CT volumes (center $=-600$ HU, width $=1500$ HU), followed by JPEG encoding with a fixed quality factor of 90, and normalizing the values to [-1,1]. The processed compressed or uncompressed CT volumes in CT-Lite, similar to CT-CLIP, are either center-cropped or zero-padded to a fixed resolution of $480 \times 480 \times 240$.  

While finetuning COLIPRI to the compressed dataset for comparison with our approach, the only difference in the preprocessing step is that the compressed volumes are resized to reach an isotropic voxel spacing of 2 mm, conforming with the original work, and the final resolution is set to $160 \times 160 \times 160$. 

\subsection{Report Preprocessing}

Radiology reports typically contain both normal and abnormal findings, with routine observations and normal impressions often constituting the majority of the text. When the entire report is tokenized and used for training, these frequent routine descriptions may dominate the textual representation and dilute the contribution of clinically meaningful tokens associated with abnormal findings.

In our framework, image–text alignment is learned using the pairwise sigmoid loss of SigLIP, which evaluates each image–text pair independently rather than relying on softmax normalization across the batch as in CLIP. Consequently, the training objective primarily benefits from sentences that describe clinically relevant abnormalities, while routine observations contribute limited supervisory signal.

To emphasize abnormality-related information, we employed MedGemma-4b-it \cite{sellergren2025medgemma} and utilize a few-shot prompting strategy to automatically filter radiology reports. The model removes sentences describing routine observations and normal findings, retaining only statements that indicate clinically significant abnormalities and impressions. This preprocessing step increases the semantic relevance of the text supervision provided to the contrastive learning objective. 

\begin{figure}[pos=h]
    \centering
    \includegraphics[width=\columnwidth]{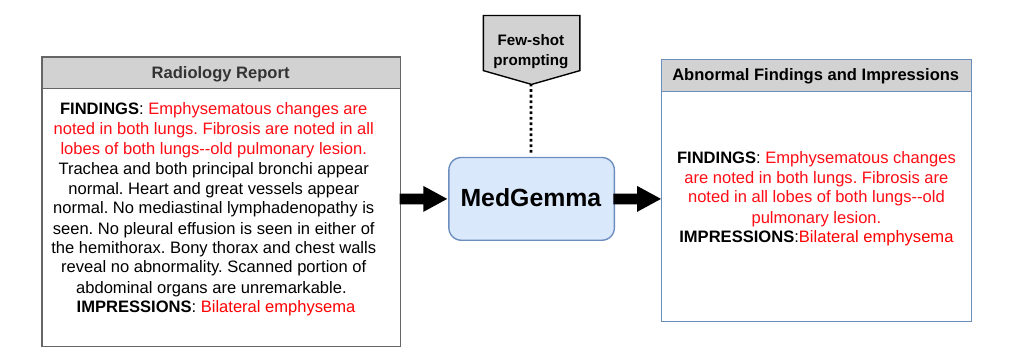}
    \caption{Filtering out abnormal findings and impressions from radiology reports via few-shot prompting}
    \label{fig:textprocessing}
\end{figure}

It is worth noting that, when fine-tuning CT-CLIP and COLIPRI for comparison with our approach, the text preprocessing procedures are kept identical to those described in the original works. Specifically, CT-CLIP utilizes full radiology reports as textual input. In contrast, COLIPRI applies few-shot prompting of a large language model to restructure the reports into eight semantic categories and to separate positive and negative findings. Maintaining these original preprocessing strategies ensures a fair comparison with the proposed method.

\subsection{Report Labeling}

We employ the same report labeler used in CT-RATE (a finetuned RadBERT-RoBERTa-4m) \cite{hamamci2026generalist} to extract thoracic abnormality labels from the radiology reports of the NIDCH dataset. The labeler is capable of extracting 18 abnormality labels. Since we strictly constrain our study to CT-volumes having undergone Lung windowing followed by JPEG-compression, we only extract 12 abnormalities that have a fair chance of getting detected by an expert under lung window setting, namely, Mosaic Attenuation Pattern, Interlobular Septal Thickening, Pulmonary Fibrotic Sequala, Lung Nodule, Peribronchial Thickening, Bronchiectasis, Emphysema, Atelectasis, Pleural Effusion, Cardiomegaly, Consolidation and Lung Opacity. The NIDCH dataset do contain adequate number of positive samples for Interlobular Septal Thickening, Peribronchial Thickening, Atelectasis and Cardiomegaly. Therefore we evaluate our models against the rest of the abnormalities. 

\subsection{Evaluation Criteria}
At the time of writing, relatively few studies have explored the use of chest CT volumes for multi-abnormality detection. Notable foundation models trained on large-scale, high-fidelity (uncompressed) CT volumes include CT-CLIP, COLIPRI, f-VLM, and CT-Net. To the best of our knowledge, prior work on multi-label thoracic abnormality prediction has primarily relied on high-fidelity CT volumes, and no previous studies have explicitly investigated learning from compressed CT volumes due to the irreversible spatial information loss introduced during compression.

In this work, we therefore evaluate CT-Lite against several existing foundation models fine-tuned on JPEG-compressed CT volumes. For CT-CLIP and COLIPRI, pretrained weights are publicly available, enabling end-to-end fine-tuning for a direct comparison with our proposed knowledge transfer approach. Since the primary objective of our study is to effectively transfer the learned knowledge of the CT-ViT visual encoder from CT-CLIP to a student encoder capable of analyzing degraded inputs, the original CT-CLIP model also serves as an important baseline. We compare the performance of CT-Lite with CT-CLIP evaluated on the same images prior to compression, as well as with CT-CLIP finetuned on the compressed dataset.

We do not include CT-Net in our experiments because the pretrained model weights are not publicly available, and pretraining the model from scratch on large-scale datasets is infeasible under our computational constraints. Similarly, f-VLM relies on anatomical masking generated by TotalSegmentator, which is designed to operate on high-fidelity NIfTI volumes. Adapting this pipeline to JPEG-compressed CT stacks would introduce substantial additional complexity. Consequently, f-VLM is also not included in our comparative evaluation.

Since we strictly constrain our study to CT-volumes having undergone Lung windowing followed by JPEG-compression, we only evaluate against 12 abnormalities that have a fair chance of getting detected by an expert under lung window setting, namely, Mosaic Attenuation Pattern, Interlobular Septal Thickening, Pulmonary Fibrotic Sequala, Lung Nodule, Peribronchial Thickening, Bronchiectasis, Emphysema, Atelectasis, Pleural Effusion, Cardiomegaly, Consolidation and Lung Opacity. However, RAD-ChestCT did not have annotations for Mosaic Attenuation Pattern, and NIDCH did not have enough positive samples for Interlobular Septal Thickening, Peribronchial Thickening, Atelectasis and Cardiomegaly. For that reason we evaluate against the rest of the abnormalities accordingly.


To evaluate the model's performance across the 18 abnormalities, we utilize Accuracy, AUROC, and the Macro-Averaged Weighted F1-Score.

\section{Result Analysis}

Table~\ref{tab:comparison} compares the performance of CT-Lite (rank = 6) with different models. Note that our primary goal is to achieve performance as close to the original CT-CLIP model as possible while inferring on compressed inputs, so the performance of CT-CLIP on the uncompressed high-fidelity counterpart of the same test dataset serves as a gold-standard performance ceiling for us. For that reason, although existing literature records some other approaches outperforming CT-CLIP, their performance on uncompressed input is not shown in this study. We show that CT-Lite significantly outperforms COLIPRI and CT-CLIP fine-tuned directly on JPEG-compressed inputs across all the datasets.

The abnormality-wise performance comparison is provided as radar plots in Supplementary Figure~\ref{fig:spiderplot}. We observe that CT-Lite performance is relatively lower for mosaic attenuation pattern, interlobular septal thickening, pulmonary fibrotic sequela, lung nodule, and peribronchial thickening, where diagnosis depends on fine details that are most affected by compression. Across the remaining abnormalities, 

\clearpage
\begin{figure}[pos=h,align=\centering]
    \begin{subfigure}{0.85\textwidth}
        \centering
        \includegraphics[width=\linewidth]{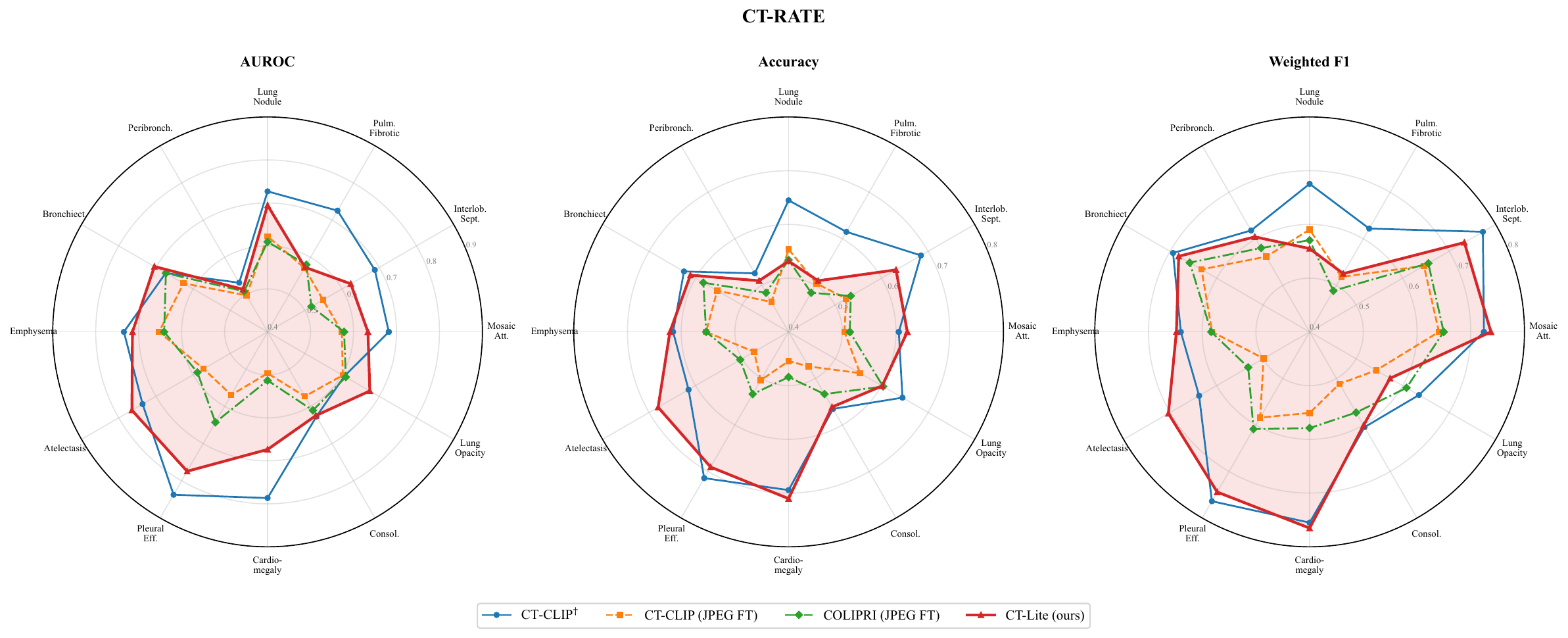}
        \caption{Performance of different approaches on CT-RATE}
        \label{fig:radar_ctrate}
    \end{subfigure}

    \vspace{0.25cm}
    \begin{subfigure}{0.85\textwidth}
        \centering
        \includegraphics[width=\linewidth]{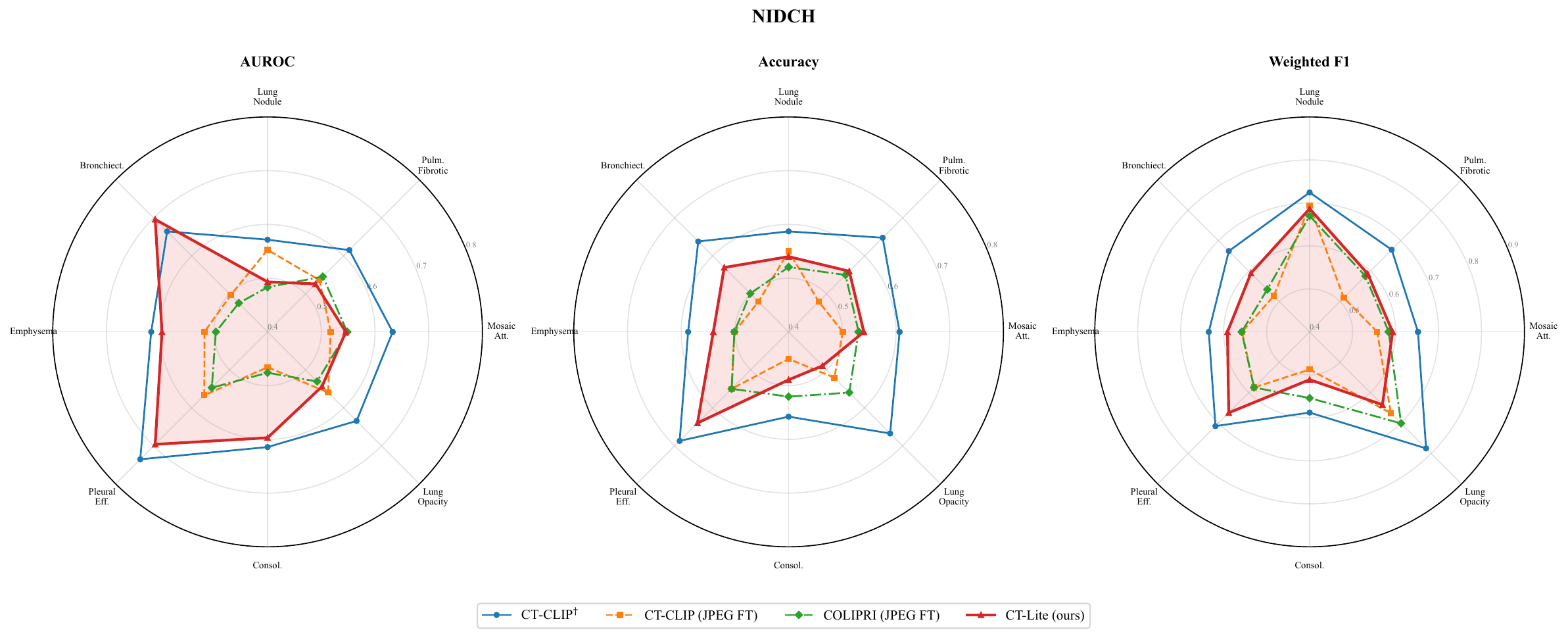}
        \caption{Performance of different approaches on NIDCH}
        \label{fig:radar_nidch}
    \end{subfigure}

    \vspace{0.25cm}
    \begin{subfigure}{0.85\textwidth}
        \centering
        \includegraphics[width=\linewidth]{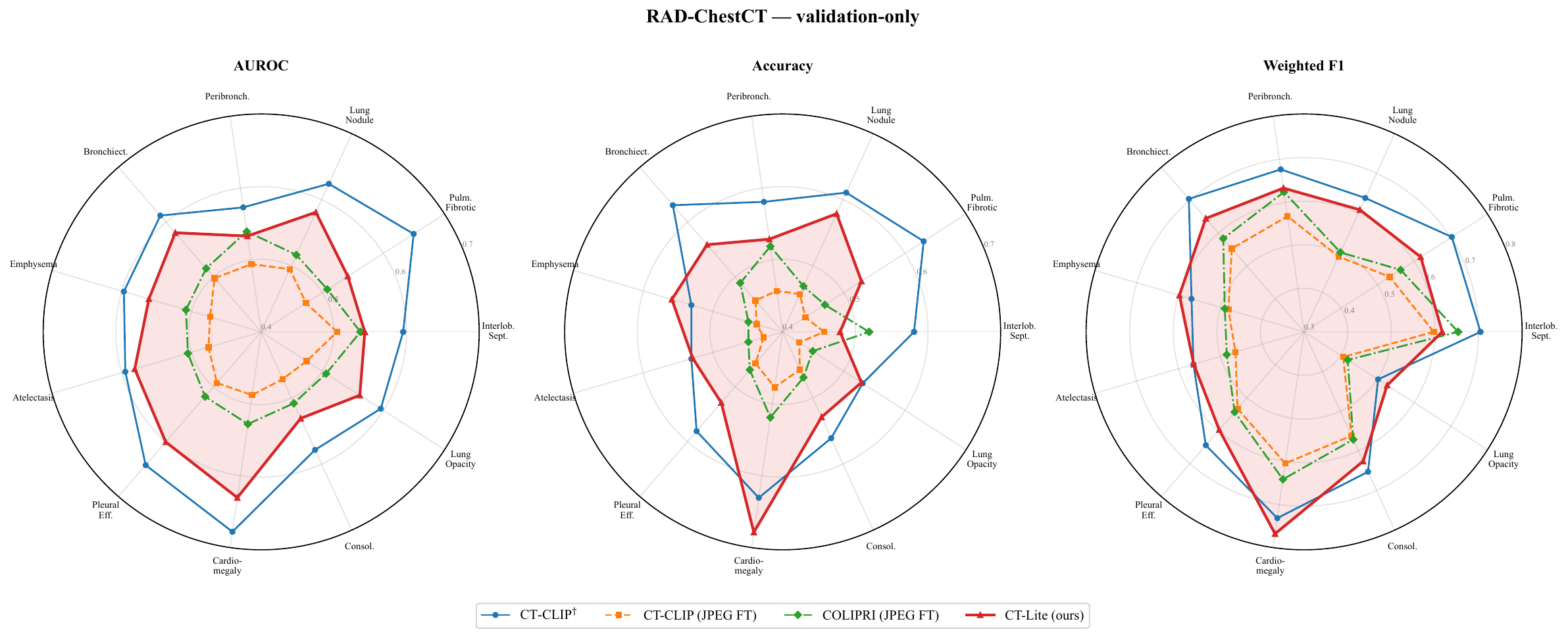}
        \caption{Performance of different approaches on RAD-ChestCT}
        \label{fig:radar_radchestct}
    \end{subfigure}
    \caption{Radar plots showing per-abnormality performance of different approaches across the three evaluation datasets. \dag\ indicates uncompressed input.}
    \label{fig:spiderplot}
\end{figure}

\clearpage

\begin{table*}[pos=h]
\centering
\caption{Comparison of Different Models on CT-RATE, NIDCH and RAD-ChestCT.}
\label{tab:comparison}
\begin{tabular}{l l c c c}
\toprule
\textbf{Dataset} & \textbf{Model} & \textbf{AUROC} & \textbf{Macro-Average} & \textbf{Accuracy} \\
 & & & \textbf{Weighted F1-Score} & \\
\midrule

\textbf{CT-RATE}
& CT-CLIP$^{\dag}$ & 0.6964 & 0.6785 & 0.6291 \\
& CT-CLIP (JPEG FT) & 0.5754 & 0.5716 & 0.5098 \\
& COLIPRI (JPEG FT) & 0.5895 & 0.5905 & 0.5288 \\
& \textbf{CT-Lite (ours)} & \textbf{0.6642} & \textbf{0.6559} & \textbf{0.6066} \\
\midrule

\textbf{NIDCH}
& CT-CLIP$^{\dag}$ & 0.6356 & 0.6786 & 0.6223  \\
& CT-CLIP (JPEG FT) & 0.5266 & 0.5721 & 0.5039 \\
& COLIPRI (JPEG FT) & 0.5128 & 0.5966 & 0.5291 \\
& \textbf{CT-Lite (ours)} & \textbf{0.5868} & \textbf{0.6090} & \textbf{0.5460}  \\
\midrule

\textbf{RAD-ChestCT}
& CT-CLIP$^{\dag}$ & 0.6128 & 0.6445 & 0.5817  \\
& CT-CLIP (JPEG FT) & 0.4855 & 0.5266 & 0.4498  \\
& COLIPRI (JPEG FT) & 0.5173 & 0.5522 & 0.4790 \\
& \textbf{CT-Lite (ours)} & \textbf{0.5677} & \textbf{0.6182} & \textbf{0.5480}  \\
\bottomrule
\end{tabular}

\vspace{2mm}
\footnotesize{$^{\dag}$Inferred on uncompressed input.}

\end{table*}

CT-Lite tracks closely with CT-CLIP on uncompressed input, suggesting that compression has little effect on these classes. Per-abnormality confusion matrices for all three datasets are provided in Supplementary Figures~\ref{fig:cm_ctrate}--\ref{fig:cm_nidch}.

\textbf{Qualitative Evaluation.}
We qualitatively evaluate the alignment of visual and text embeddings of CT-Lite using t-Stochastic Neighbor Embedding (t-SNE) visualizations (Supplementary Figure~\ref{fig:tsne}); both visual embeddings from input CT volumes and textual embeddings from the corresponding radiology reports are projected into a 2D space using all validation data. The separation between positive and negative contours is consistent with the per-abnormality performance reported in Supplementary Figure~\ref{fig:spiderplot}, and the similarity between the visual and textual embedding contours indicates successful multimodal alignment.

\begin{figure*}[pos=h]
    \centering
    \includegraphics[width=\textwidth]{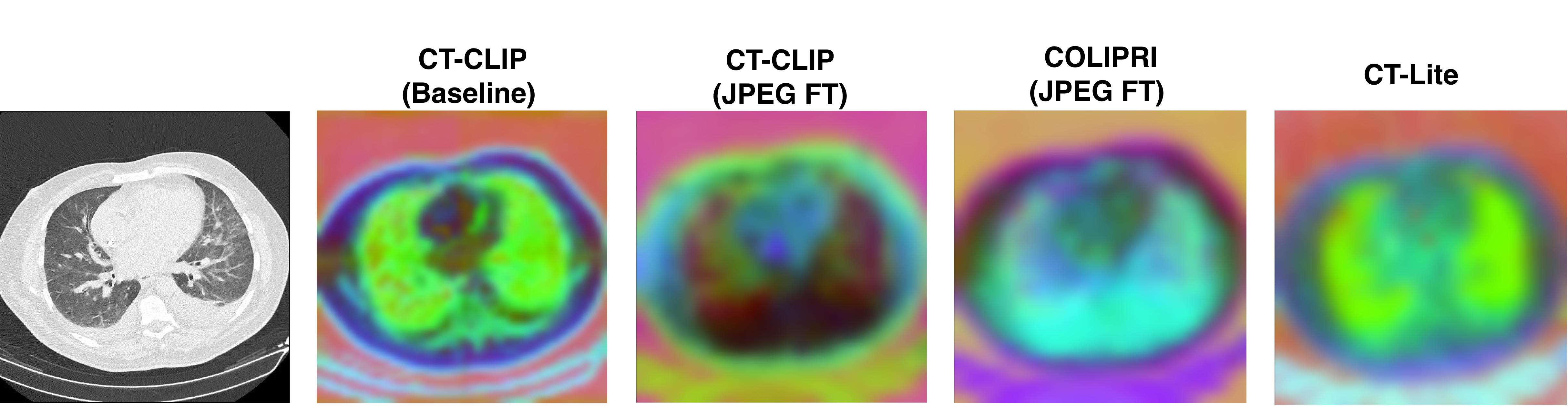}
    \caption{Principal Component Analysis maps of the visual encoder output vectors for one chest CT slice. Compressed CT (column 1): the JPEG-compressed input. CT-CLIP (Baseline) (column 2): inferred on the uncompressed counterpart of the same slice. Remaining columns: all inferred on the JPEG-compressed input. Visual embeddings are upsampled with bicubic interpolation.}
    \label{fig:pca}
\end{figure*}

We also visualize Principal Component Analysis (PCA) maps of the different visual encoders inferred on CT images, as shown in Figure~\ref{fig:pca}. Following \cite{wald2026_colipri}, the visual embeddings are upsampled with bicubic interpolation for visualization. The original CT-CLIP is inferred on an uncompressed input, while the remaining columns are inferred on the JPEG-compressed counterpart. CT-CLIP on high-fidelity input produces a sharp feature map, whereas CT-CLIP and COLIPRI-CRM fine-tuned on compressed data fail to recover inherent details, yielding blurry and noisy PCA maps. CT-Lite, in contrast, produces a PCA map closely resembling CT-CLIP's despite operating on compressed input, with only minor artifacts attributable to the irreversible loss of spatial information caused by compression.

\begin{figure}
    \centering
    \includegraphics[width=\columnwidth]{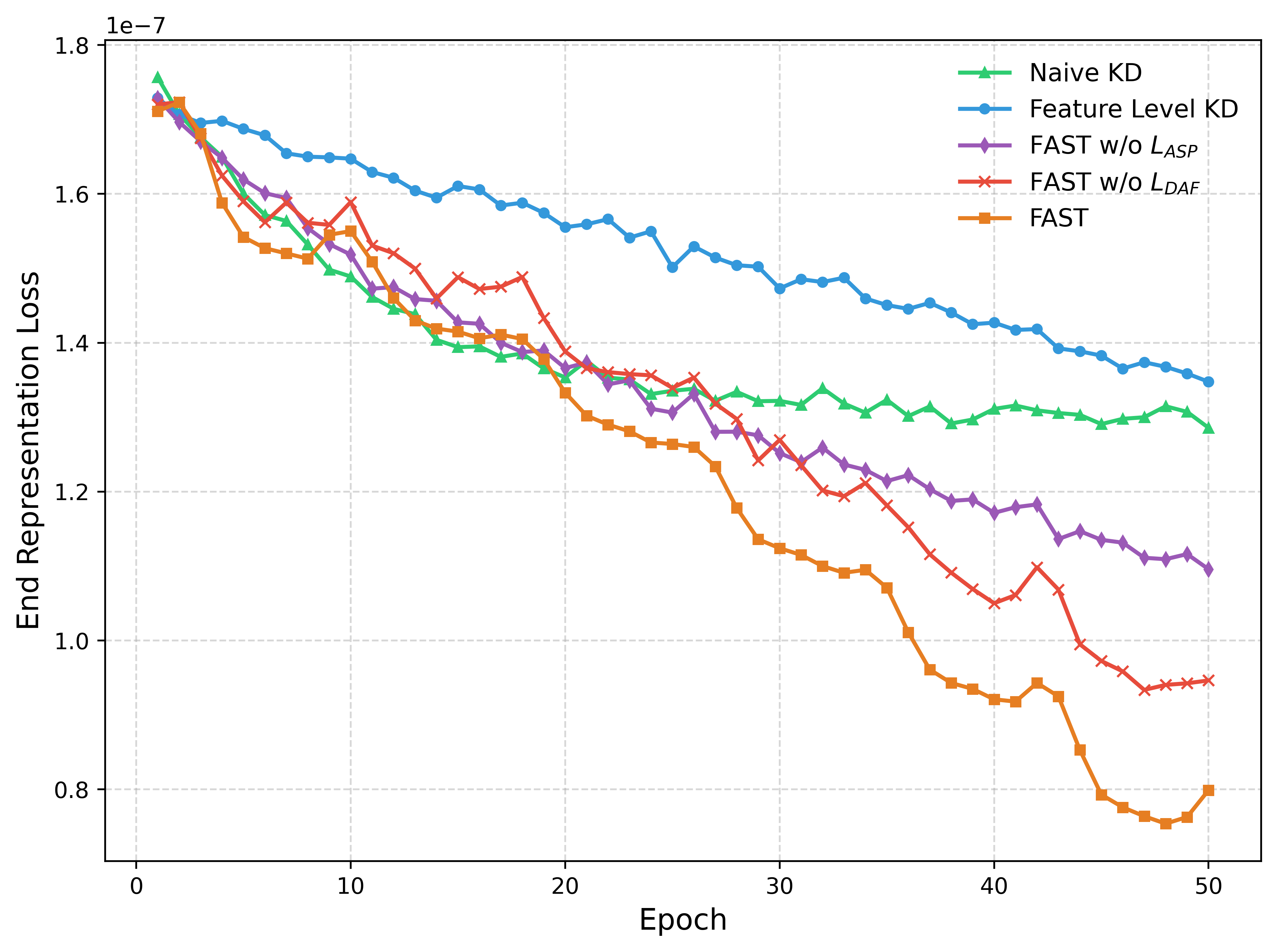}
    \caption{End-representation MSE loss for five distillation strategies on a 50-volume held-out set: naive KD, feature-level KD, FAST w/o $\mathcal{L}_{\text{ASP}}$, FAST w/o $\mathcal{L}_{\text{DAF}}$, and full FAST.}
    \label{fig:ablation_convergence}
\end{figure}

\section{Ablation Studies}

To validate the effectiveness of our proposed components, we conducted ablation studies to quantify the contribution of each loss component in the Feature Attention Style Transfer (FAST) framework. To determine the optimal loss function, five independent experiments are carried out for 50 epochs on a small dataset of 50 CT volumes, using AdamW with a cosine annealing scheduler and a learning rate of $1 \times 10^{-5}$ across all five runs. As shown in Figure~\ref{fig:ablation_convergence}, the full FAST framework achieves the fastest convergence and the lowest end-representation loss compared with naive knowledge distillation (KD), feature-level KD, and partial variants that omit either $\mathcal{L}_{ASP}$ or $\mathcal{L}_{DAF}$. This demonstrates the synergistic effect of combining attention style preservation with dual-attention feature alignment: removing any single component leads to a clear degradation in the convergence behavior.

\section{Scaling Laws}

\begin{figure*}[pos=h]
    \centering
    \begin{subfigure}{0.32\textwidth}
        \includegraphics[width=\linewidth]{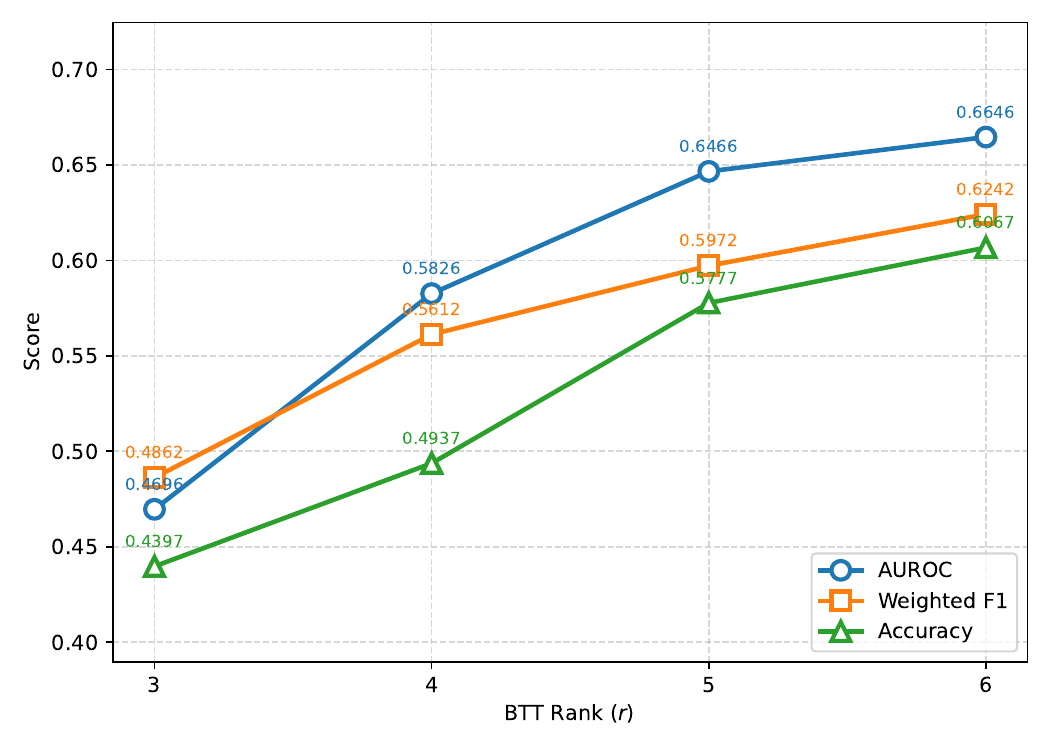}
        \caption{CT-RATE}
        \label{fig:rank_sweep_ctrate}
    \end{subfigure}
    \hfill
    \begin{subfigure}{0.32\textwidth}
        \includegraphics[width=\linewidth]{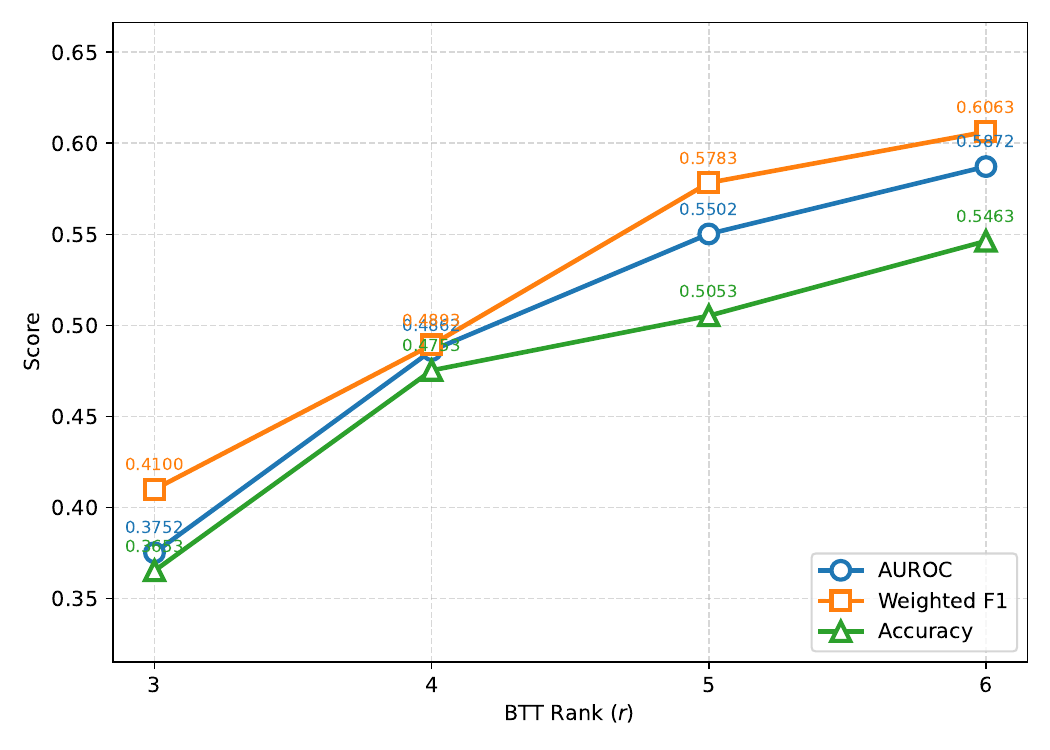}
        \caption{NIDCH}
        \label{fig:rank_sweep_nidch}
    \end{subfigure}
    \hfill
    \begin{subfigure}{0.32\textwidth}
        \includegraphics[width=\linewidth]{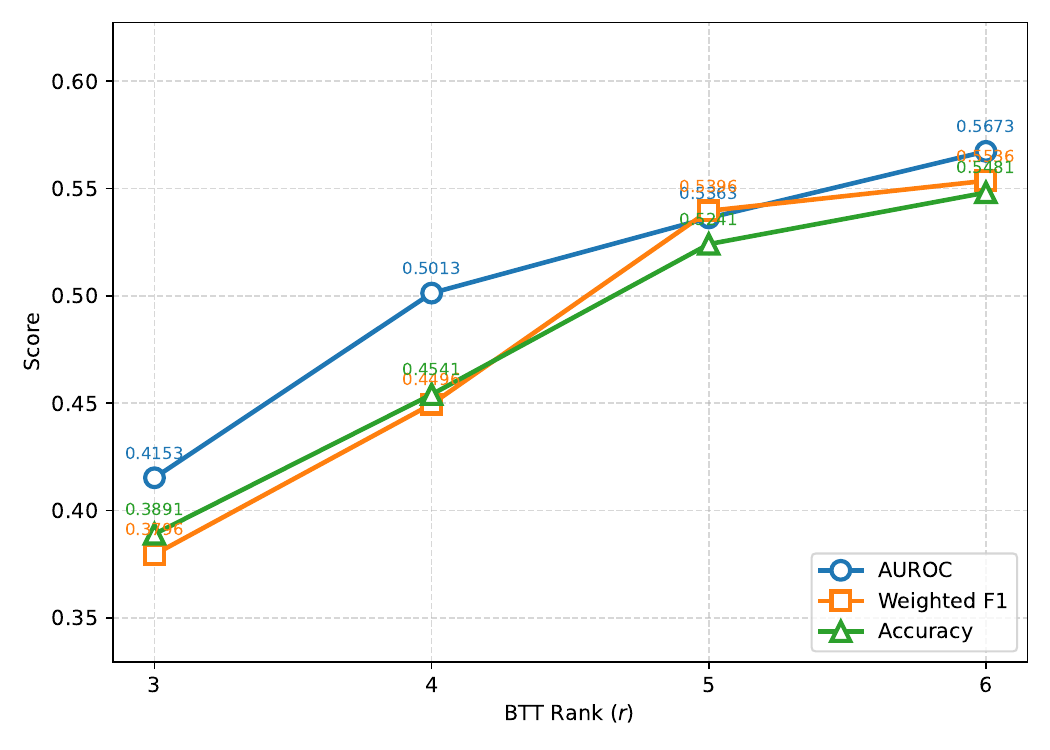}
        \caption{RAD-ChestCT}
        \label{fig:rank_sweep_radchestct}
    \end{subfigure}
    \caption{AUROC (blue), Weighted F1 (orange), and Accuracy (green) of CT-Lite across Block Tensor Train ranks $r \in \{3, 4, 5, 6\}$ on (a) CT-RATE, (b) NIDCH, and (c) RAD-ChestCT. All metrics increase monotonically with rank across the three datasets, with $r = 6$ yielding the best performance.}
    \label{fig:rank_sweep}
\end{figure*}

We also evaluate the impact of the rank parameter in our Structured Factorized Projection (SFP) layer. As derived in Section~III, for $\text{out\_dim} = 512$ the efficiency bound requires $r \leq \sqrt{512}/2 \approx 11.3$. We sweep over ranks $r \in \{3, 4, 5, 6\}$ as permitted by our resource budget, all of which satisfy the bound and guarantee parameter savings over the dense baseline. Figure~\ref{fig:rank_sweep} reports the trade-off between projection-head capacity and performance: as expected, all three metrics increase monotonically with rank, and $r = 6$ achieves the best results across all three datasets while remaining within the efficiency bound. Normalized against CT-CLIP's dense visual projection, the relative floating-point operations (FLOPs) for SFPs with ranks $6,5,4,3$ are $0.53$, $0.44$, $0.35$, and $0.27$, respectively.

\section{Conclusion and Future Works}

In this work, we present CT-Lite, a comprehensive framework for compute-efficient multimodal chest CT analysis from JPEG-compressed volumes. The Feature Attention Style Transfer (FAST) framework distills both activation patterns and structural relationships from a high-fidelity spatiotemporal teacher to a student operating on compressed inputs, while Structured Factorized Projection (SFP) replaces the dense projection head with a Block Tensor Train decomposition, cutting projection parameters from approximately 1.07~B to 569~M and lowering relative projection FLOPs to 0.27--0.53 across the rank sweep. Despite operating on degraded inputs and with substantially fewer parameters, CT-Lite stays within roughly 5--7\% AUROC of the uncompressed-input CT-CLIP baseline on the in-domain CT-RATE test set and on the external NIDCH and RAD-ChestCT cohorts, the latter two acquired on different scanners and at different institutions than the training data.

Several limitations remain. A measurable AUROC gap to the uncompressed baseline persists across all three datasets, the contrastive pretraining corpus is modest in size, and the framework is not yet validated prospectively in a clinical workflow; CT-Lite is therefore not yet at the level required for deployment in clinical settings.

We see several natural directions for future work. First, additional studies are needed to close the residual performance gap between compressed-input and uncompressed-input pipelines, for example through stronger compression-aware distillation objectives, larger student capacities, or codec-aware pretraining. Second, more effective compression strategies for volumetric medical imaging are worth exploring beyond standard JPEG, including JPEG-XL, learned codecs, and task-aware compression. Finally, we believe the broader question of leveraging compressed inputs for volumetric medical imaging is largely open, and we hope this work motivates further investigation into compressed-domain learning for other anatomies and modalities.

\section{Acknowledgements}
This work is supported by Research and Innovation Center for Science and Engineering (RISE), Bangladesh University of Engineering and Technology (BUET), under Grant No. S2024-01-012. We extend our gratitude to Dr. Sabina Akhter, Associate Professor, and Dr. Fahmida Sharmin, Radiologist, from the Department of Radiology and Imaging, National Institute of Diseases of the Chest and Hospital (NIDCH), Dhaka, Bangladesh, for their invaluable assistance in data collection procedure. We also thank Noushin Yousuf Shanin, and M.M. Nayeem Hasan, for their assistance in data curation process.  

\section{Ethical Statement}

CT-RATE is a publicly available dataset. The NIDCH dataset is collected after ethical approval from the Director of the National Institute of Diseases of the Chest and Hospital (NIDCH) on 27 November 2023, after the data had been completely anonymized and de-identified by removing all personally identifiable information (PII) and protected health information (PHI). The dataset is intended for research purposes only and should not be used for commercial applications without appropriate permissions. Samples of RAD-ChestCT are accessed under an open license under the Creative Commons Attribution-NonCommercial-NoDerivatives 4.0 International (CC BY-NC-ND 4.0).

\FloatBarrier 
\bibliographystyle{elsarticle-num}
\bibliography{references}

\clearpage
\onecolumn
\noindent{\Large\bfseries Supplementary Figures}\par\medskip
\renewcommand{\thefigure}{S\arabic{figure}}
\setcounter{figure}{0}




\begin{figure}[pos=h]
    \centering
    \includegraphics[width=\textwidth,height=0.81\textheight]{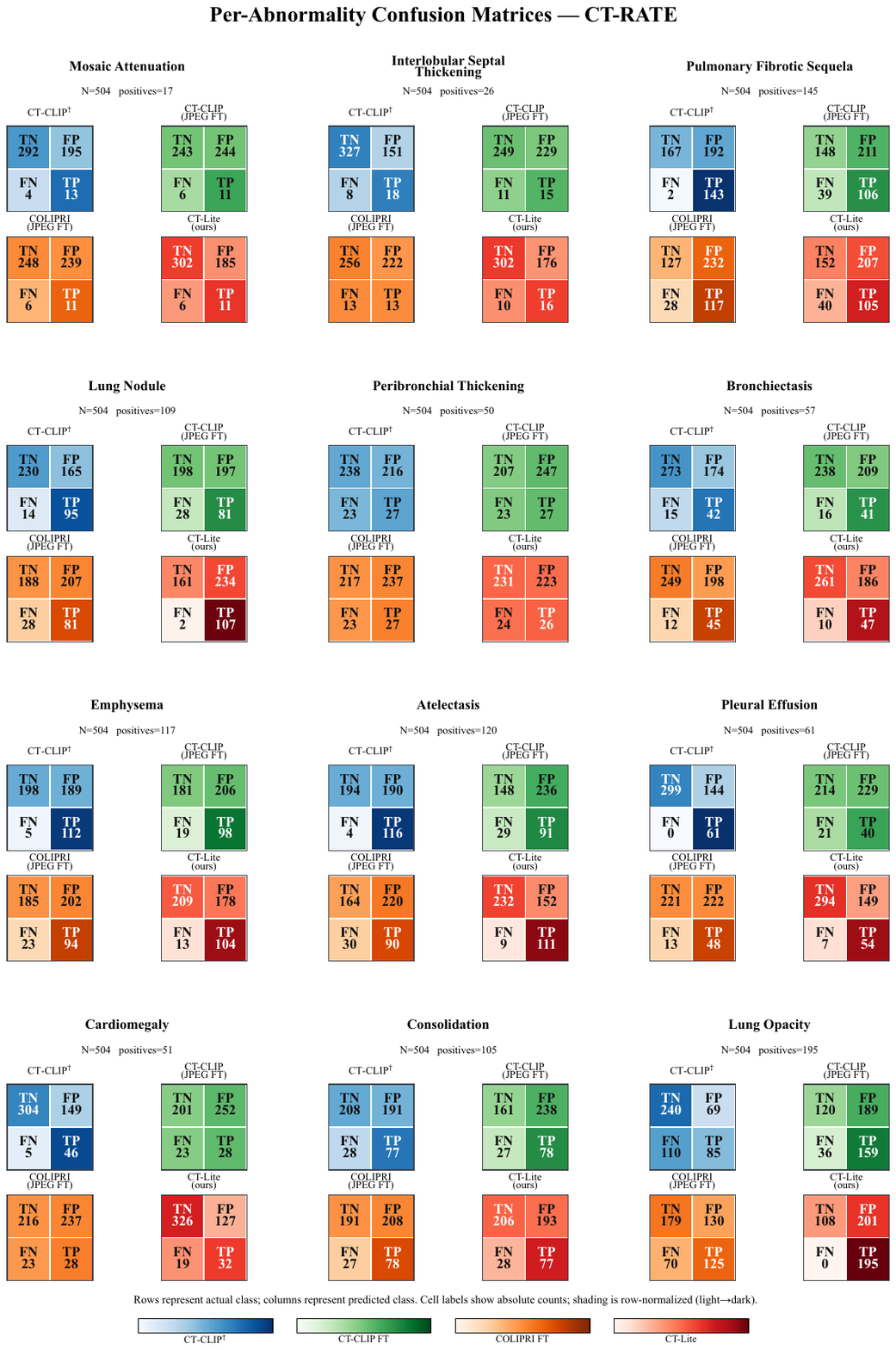}
    \caption{Per-abnormality confusion matrices on the CT-RATE dataset for CT-CLIP\textsuperscript{\dag}, CT-CLIP (JPEG FT), COLIPRI (JPEG FT), and CT-Lite (ours). Each panel corresponds to one abnormality; within each panel, the four confusion matrices compare the models. Cells are row-normalized for color and annotated with absolute counts. \dag\ indicates uncompressed input.}
    \label{fig:cm_ctrate}
\end{figure}

\begin{figure}[pos=h]
    \centering
    \includegraphics[width=\textwidth,height=0.95\textheight]{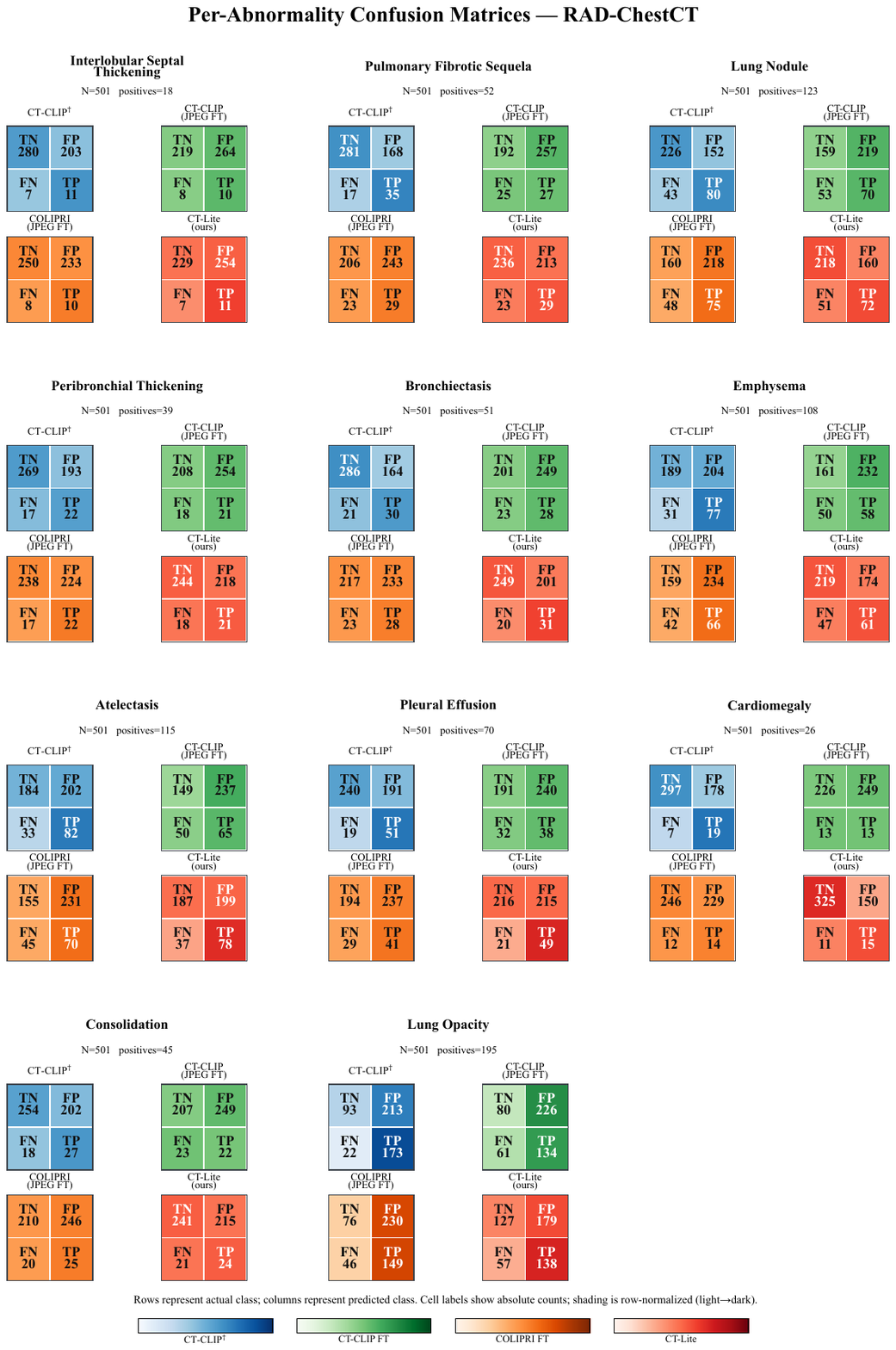}
    \caption{Per-abnormality confusion matrices on the RAD-ChestCT dataset for CT-CLIP\textsuperscript{\dag}, CT-CLIP (JPEG FT), COLIPRI (JPEG FT), and CT-Lite (ours). Each panel corresponds to one abnormality; within each panel, the four confusion matrices compare the models. Cells are row-normalized for color and annotated with absolute counts. \dag\ indicates uncompressed input.}
    \label{fig:cm_radchestct}
\end{figure}

\begin{figure}[pos=h]
    \centering
    \includegraphics[width=\textwidth,height=0.95\textheight]{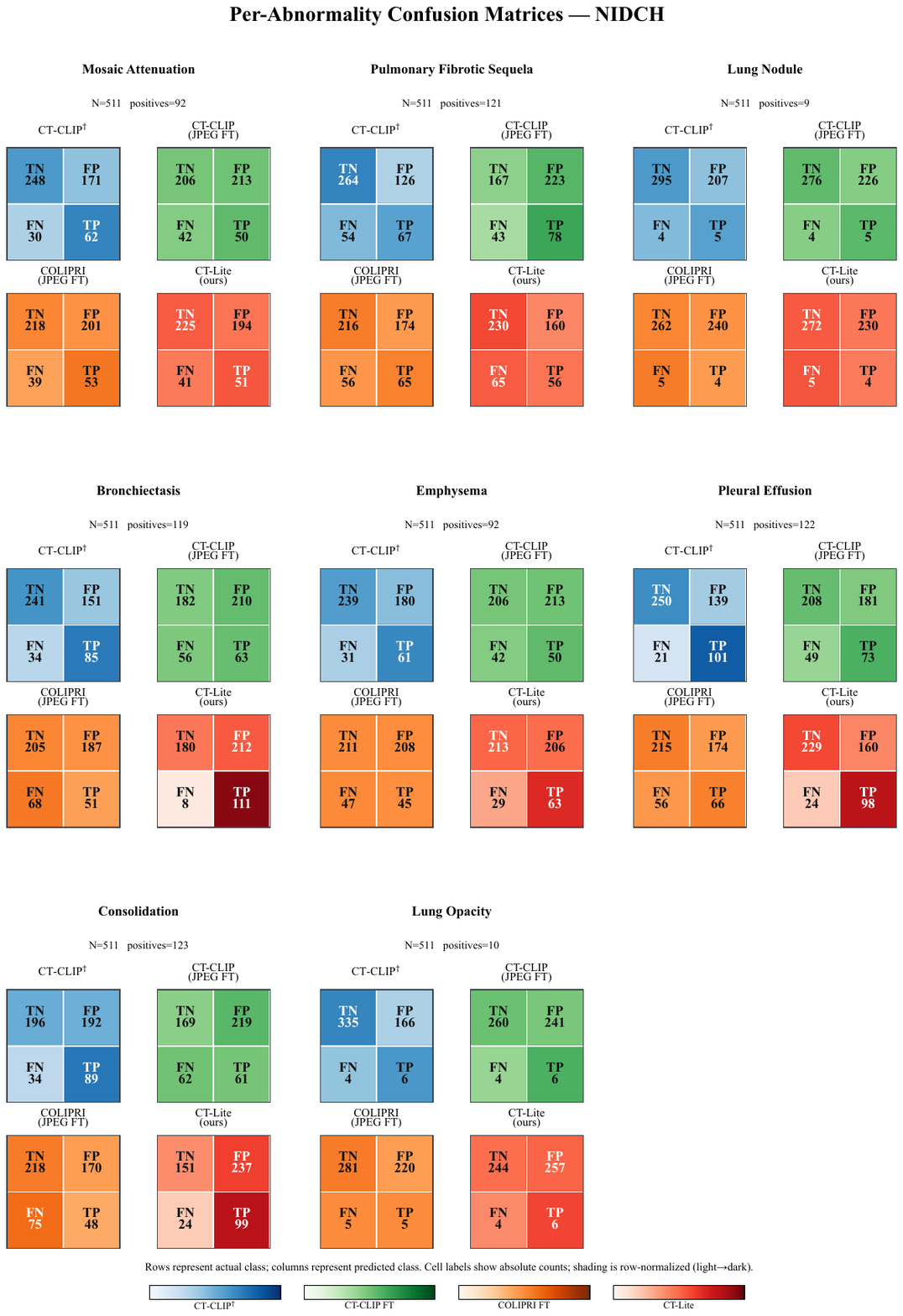}
    \caption{Per-abnormality confusion matrices on the NIDCH dataset for CT-CLIP\textsuperscript{\dag}, CT-CLIP (JPEG FT), COLIPRI (JPEG FT), and CT-Lite (ours). Each panel corresponds to one abnormality; within each panel, the four confusion matrices compare the models. Cells are row-normalized for color and annotated with absolute counts. \dag\ indicates uncompressed input.}
    \label{fig:cm_nidch}
\end{figure}

\begin{figure}[!htbp]
    \centering
    \includegraphics[height=0.85\textheight]{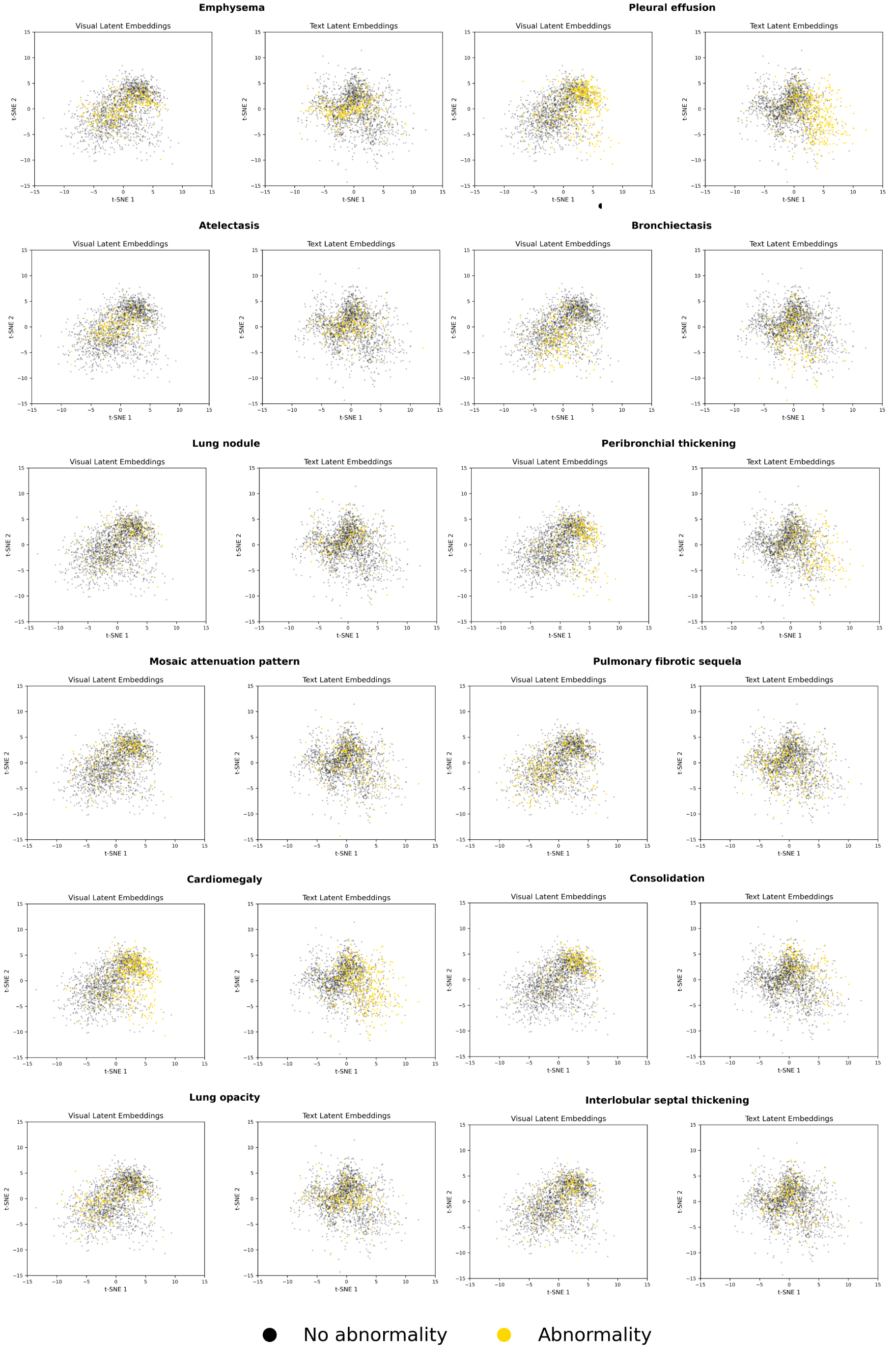}
    \caption{t-SNE projections of CT-Lite visual embeddings (from input CT volumes) and textual embeddings (from corresponding radiology reports), using all validation data. Black markers denote samples without a given abnormality and yellow markers denote samples with the abnormality.}
    \label{fig:tsne}
\end{figure}


\end{document}